\ifdefined\pdfminorversion\pdfminorversion=7\fi
\PassOptionsToPackage{numbers,sort&compress,square}{natbib}
\documentclass[11pt,a4paper,onecolumn]{thuc3i}
\usepackage{custom}

\usepackage{xspace}
\usepackage{xcolor}
\usepackage{tcolorbox}
\usepackage{booktabs}
\usepackage{longtable}
\usepackage{graphicx}
\usepackage{colortbl}
\usepackage{multirow}
\usepackage{etoolbox}
\usepackage{caption}
\tcbuselibrary{skins,breakable,listings}
\usepackage[table]{xcolor}
\definecolor{groupblue}{RGB}{235,243,252}
\definecolor{bestgreen}{RGB}{225,244,229}
\definecolor{obsTitleBg}{HTML}{D9D9D9}
\definecolor{obsBodyBg}{HTML}{F6F6F6}
\definecolor{obsBorder}{HTML}{DADADA}
\definecolor{obsLeftBar}{HTML}{777777}

\newtcolorbox{observationbox}[1][Observation]{
    enhanced,
    breakable,
    width=\linewidth,
    title={#1},
    fonttitle=\bfseries,
    coltitle=black,
    colbacktitle=obsTitleBg,
    colback=obsBodyBg,
    colframe=obsBorder,
    boxrule=0.4pt,
    titlerule=0pt,
    arc=2mm,
    sharp corners=west,
    borderline west={2pt}{0pt}{obsLeftBar},
    boxsep=0pt,
    left=3mm,
    right=3mm,
    top=2mm,
    bottom=2mm,
    toptitle=1mm,
    bottomtitle=1mm,
    before skip=8pt,
    after skip=8pt
}

\newtcolorbox{insightbox}[1][Insights for Workflow Design]{
    enhanced,
    breakable,
    width=\linewidth,
    title={#1},
    fonttitle=\bfseries,
    coltitle=black,
    colbacktitle=obsTitleBg,
    colback=obsBodyBg,
    colframe=obsBorder,
    boxrule=0.4pt,
    titlerule=0pt,
    arc=2mm,
    sharp corners=west,
    borderline west={2pt}{0pt}{obsLeftBar},
    boxsep=0pt,
    left=3mm,
    right=3mm,
    top=2mm,
    bottom=2mm,
    toptitle=1mm,
    bottomtitle=1mm,
    before skip=8pt,
    after skip=8pt
}

\definecolor{HEBlue}{HTML}{357ABD}
\definecolor{HERed}{HTML}{D95F5F}

\newcommand{\HEgain}[2]{%
    \cellcolor{HEBlue!#1!white}\color{black}#2}
\newcommand{\HEnongain}[2]{%
    \cellcolor{HERed!#1!white}\color{black}#2}
\newcommand{\name}{Coda-DaVinci\xspace}

\setheadertext{\name}
\title{\name: Reimagining Video Generation with Code2Video}

\author[1,2,*]{Yiyan Ji}
\author[3,2,*]{Qixun Wang}
\author[3,2,*,$\spadesuit$]{Yang Shi}
\author[3]{Zhuoran Zhang}
\author[1]{Yifan Xu}
\author[3]{Tengfei Liu}
\author[4]{Yuqi Tang}
\author[1]{Xiaohan Zhang}
\author[5]{Shujie Li}
\author[1]{Jiafu Tang}
\author[3]{Xuanyu Zhu}
\author[6]{Xinyu Liu}
\author[3]{Yan He}
\author[2]{Yushuo Guan}
\author[2]{Zihao Wang}
\author[3,$\dagger$]{Xianghua Ying}
\author[7]{Shixuan Liu}
\author[7,$\dagger$]{Haotian Wang}
\makeatletter
\begingroup
\renewcommand{\AB@authnote}[1]{}
\author[0]{Wenjing Yang}
\endgroup
\makeatother
\author[1]{Zhaoxiang Zhang}
\author[2,$\spadesuit$]{Yuanxing Zhang}
\author[2]{Pengfei Wan}
\author[1,$\dagger$]{Jiaheng Liu}

\affil[1]{Nanjing University}
\affil[2]{KlingAI Research}
\affil[3]{Peking University}
\affil[4]{Hong Kong University of Science and Technology (Guangzhou)}
\affil[5]{University of Hong Kong}
\affil[6]{Hong Kong University of Science and Technology}
\affil[7]{Tsinghua University}

\role[*]{Equal Contribution}
\role[\dagger]{Corresponding Author}
\role[\spadesuit]{Project Lead}

\resource{}{%
  \href{https://coda-davinci.github.io/}{\faHome~Homepage}%
  \hspace{1.5em}%
  \href{https://github.com/FrankYang-17/Coda}{\faGithub~Code}%
}

\begin{abstract}
Code2Video uses agent-authored 3D proxies to guide video generation, enabling explicit control over structure and dynamics. However, how to construct proxies and prompts for optimal downstream quality remains underexplored, as do the capabilities and limitations of this paradigm. Through controlled experiments and blinded human evaluation, we study proxy granularity, physics simulation, and prompting. The results reveal task-dependent trade-offs: coarse proxies favor natural motion, while medium-grained proxies favor consistency. Physics simulation and proxy-aligned prompting further improve average quality. Guided by these findings, we introduce \textbf{\name}, a task-adaptive agentic framework for Code2Video generation that selects between proxy-guided and direct generation, adapts modeling granularity to each entity, and coordinates simulation with proxy-grounded prompting. Its multi-agent harness preserves scene assets and guides repairs through visual inspection and independent review. Experiments demonstrate leading instruction following on VTR and scene consistency on R2M, improved physical plausibility over direct generation, and higher visual realism than the compared Code2Video baselines. Long-video evaluations further show gains in event structure and consistency across successive interactions, while revealing limitations in transition stability and adherence to scene settings.
\end{abstract}

\makeatletter
\renewcommand{\abscontent}{%
    \par\noindent
    \textbf{Abstract $\vert$}\enspace
    {\absfont\theabstract\par}
    \@ifundefined{@keywords}{}{%
        \vskip1em
        \noindent\keywordsfont Keywords: \@keywords
    }%
}

\patchcmd{\maketitle}
    {\vskip8pt}
    {%
        \vskip8pt
        \begin{center}
            \includegraphics[
                width=\textwidth,
                keepaspectratio
            ]{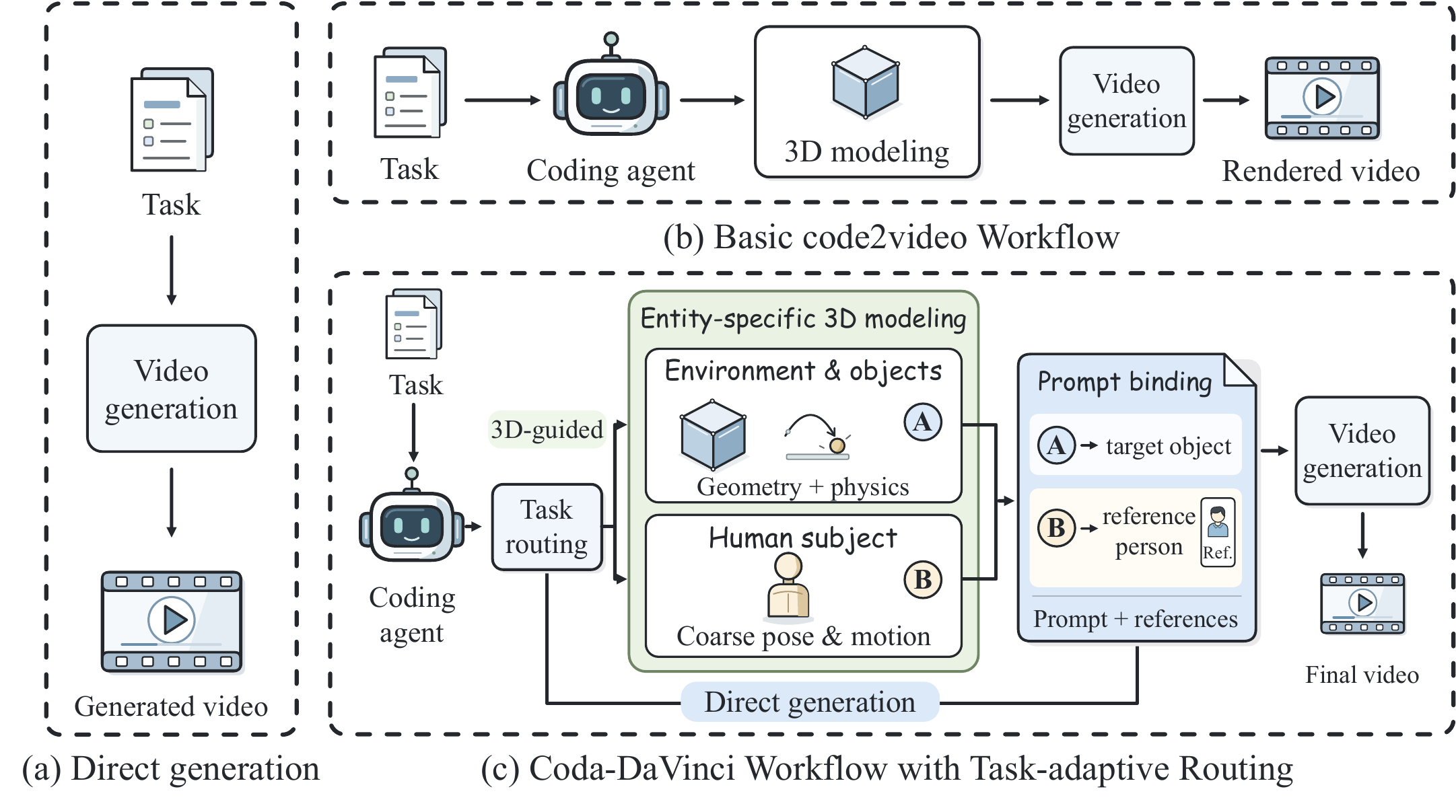}
            \captionsetup{
                font=small,
                skip=4pt,
                justification=raggedright,
                singlelinecheck=false
            }
            \captionof{figure}{%
                \textbf{Comparison of video generation workflows.}
                (a) Direct video generation.
                (b) Basic Code2Video workflows model the inputs as 3D scenes and render them into the final video.
                (c) \name combines task-level routing with multi-granularity 3D modeling and generates prompts for subject binding.
            }
            \label{fig:comparison}
        \end{center}
        \vspace{4pt}
    }
    {}
    {\PackageError{teaser}{Failed to insert teaser}{}}
\makeatother

\begin{document}
\maketitle
\section{Introduction}
\label{sec:intro}

\begin{figure*}[htbp]
    \centering
    \includegraphics[width=\textwidth]{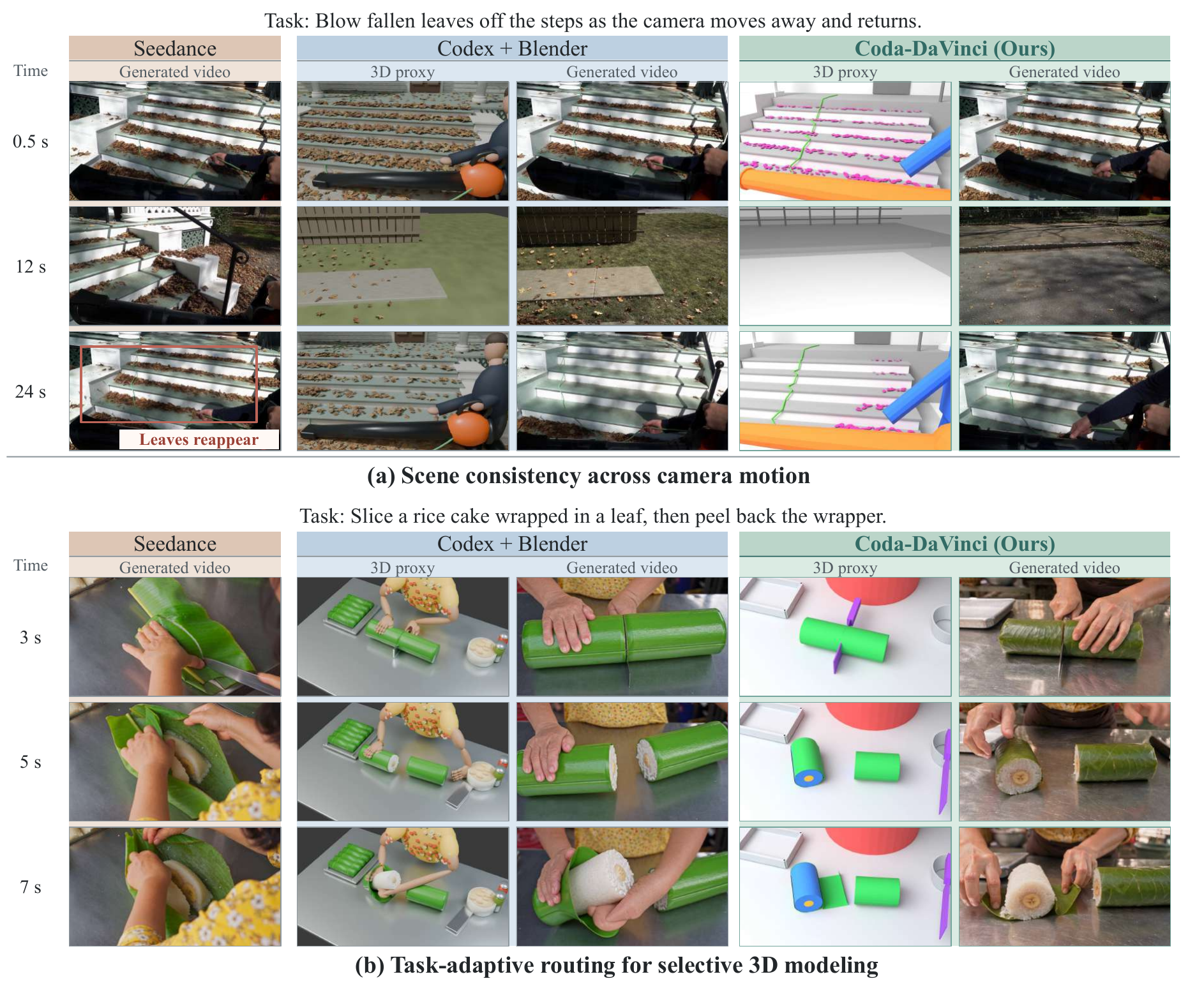}
    \caption{\textbf{Selective 3D modeling.} Each method pairs 3D proxy frames with generated video frames at 3, 5, and 7 seconds. Codex + Blender and VideoCoCo explicitly model hands and arms. \name models the objects involved in cutting and unwrapping, while leaving detailed hand articulation to the video generator.}
    \label{fig:teaser}
\end{figure*}

Recent video generation models can synthesize realistic and visually appealing videos from text and reference media~\cite{seedance2026seedance,Kling,wan2025,LTX-2,tang2026codecap}.
Yet following detailed instructions, preserving scene state over time, and producing physically plausible interactions remain challenging.
Code2Video approaches, including VideoCoCo~\cite{li2026videococo} and Code Video Model~\cite{liang2026codevideomodel}, address these challenges through executable scene programs.
In this paradigm, a coding agent constructs and renders a 3D proxy whose structure and motion guide a video model in synthesizing the final video. This explicit scene representation helps maintain scene state across camera revisits, as shown in Figure~\ref{fig:teaser}(a).

The benefits of Code2Video depend on the task and on how the proxy is constructed and used.
Scene programs provide precise control over layout and trajectories, but detailed modeling can introduce geometric or motion errors that may carry over to the final video.
When motion naturalness is central and explicit scene control offers limited benefit, an imperfect proxy may yield less natural results than direct generation.
Proxies for the same task can differ substantially in their level of detail.
Physical interactions also require choosing between scripted motion and physics simulation.
The synthesis prompt must then align the intended appearance and finer actions with the rendered proxy.
These considerations raise two central questions:
\emph{How can we achieve the best video generation performance within the Code2Video paradigm, and how far can Code2Video extend the capabilities of video generation?}


We study these questions through controlled comparisons with blinded human evaluation on tasks involving scene revisitation, complex motion, and physical processes.
The benefits of proxy guidance vary by task: improved consistency and instruction following can come at the cost of motion naturalness.
Under matched settings, coarse proxies preserve more natural motion than detailed alternatives on motion-intensive tasks, while medium-grained proxies provide stronger consistency when physics simulation is required.
Prompts written by the coding agent outperform the tested external enhancement strategy in pooled scores.
Among the tested configurations, medium-grained proxies combined with physics simulation and agent-authored prompts achieve the highest pooled performance.

Guided by these findings, we introduce \textbf{\name}, a task-adaptive agentic framework for Code2Video generation as shown in Figure~\ref{fig:comparison}.
\name selects 3D modeling when precise scene or motion control is needed and favors direct generation for natural body movement when explicit control is unnecessary and proxy animation risks stiffness.
For modeled scenes, it uses medium-grained proxies for environments and structured objects, coarse proxies for human subjects, and Blender simulation for supported physical interactions, as shown in Figure~\ref{fig:teaser}(b).
The agent inspects the proxy before writing a synthesis prompt that links appearance references to proxy entities and specifies finer human actions.
A multi-agent harness preserves the scene workspace across revisions and successive requests, supports visual inspection, and uses an independent acceptance agent to identify unmet requirements for repair.

Experiments show that \name leads the compared methods in instruction following on VTR~\cite{huang2026vtr} and scene consistency on R2M~\cite{gu2026r2m}.
It also improves physical plausibility over direct generation on World Models' Last Exam in Physics~\cite{gao2026lastexam} and outperforms Code2Video baselines in visual realism on the T2AV-Compass subset~\cite{cao2026t2avcompass}, approaching direct generation.
We also examine agent--generator pairings and rendering engines; Blender offers higher quality than Three.js at greater token cost.
Long-video and multi-turn evaluations show gains in long-form structure and scene consistency over direct continuation, but lower transition stability and setting adherence.

Our main contributions are threefold:
\begin{itemize}
    \item We systematically analyze proxy granularity, physics simulation, and prompting, deriving task-dependent guidelines for proxy construction and video synthesis.
    \item We introduce \textbf{\name}, a task-adaptive agentic framework that coordinates selective 3D modeling and proxy-grounded synthesis, supported by a persistent scene workspace and visual feedback.
    \item We empirically characterize Code2Video's capabilities and limitations across tasks, agent--generator pairings, rendering engines, and long-video settings, revealing trade-offs in visual quality, consistency, and token cost.
\end{itemize}

\section{Related Work}
\label{sec:related_work}

\textbf{Video Generation.}
Recent advances in video generation have substantially improved visual fidelity, motion quality, and text--video alignment~\cite{seedance2026seedance,Kling,wan2025,LTX-2,kim2026rethinking,wu2026reprompt,shi2025mavors,team2026behind,tang2026keyframe,liu2026longav,zhang2026multiref}.
Building on this progress, recent work increasingly emphasizes reliable instruction following, spatiotemporal consistency, and physical plausibility.
TempAct~\cite{wang2026tempact} improves adherence to temporally complex instructions through planner--executor reinforcement learning.
To maintain visual consistency over longer sequences, Context Forcing~\cite{chen2026context} combines long-context supervision with slow--fast memory, while LongLive-RAG~\cite{hu2026longlive} retrieves previously generated video latents to mitigate identity drift.
MMPhysVideo~\cite{lin2026mmphysvideo} improves physical plausibility by jointly modeling RGB video with semantic, geometric, and trajectory cues.
Despite these advances, reliably following complex instructions, maintaining spatiotemporal consistency, and generating physically plausible dynamics remain challenging.

\textbf{Code2Video.}
Code2Video generation uses executable programs to control visual content and motion, enabling targeted editing.
The Code2Video framework~\cite{chen2026code2video} targets educational videos, and MoVer~\cite{ma2025mover} guides motion graphics refinement through spatiotemporal verification.
StoryBlender~\cite{li2026storyblender} produces consistent 3D storyboards using persistent scene memory and engine-based verification, while SimWorlds~\cite{liu2026simworlds} constructs dynamic Blender scenes through planning, coding, and verification.
For proxy-guided synthesis, GPT4Motion~\cite{lv2024gpt4motion} uses Blender-simulated motion cues to guide diffusion models.
VideoCoCo~\cite{li2026videococo} guides video generation with Blender-rendered spatiotemporal drafts, while Code Video Model~\cite{liang2026codevideomodel} combines Three.js proxies with early-step conditioning to preserve structure and reduce synthetic-appearance leakage.

\section{Empirical Analysis}
\label{sec:empirical_analysis}

\subsection{Code2Video Pipeline Formulation}
Given an input prompt $x$ with optional image or video references $r$, a coding agent $\mathcal{A}$ constructs a proxy video $\tilde{v}$. A video-generation model $\mathcal{G}$ then produces the final video $v$ conditioned on the proxy and a generation prompt $q$:
\begin{equation}
\tilde{v}=\mathcal{A}(x,r),
\qquad
v=\mathcal{G}(\tilde{v},q,r),
\end{equation}
where $q$ is either the original prompt $x$ or an enhanced version produced by the coding agent $\mathcal{A}$ or an external model.

To address the central questions raised in Section~\ref{sec:intro}, we examine how different proxy construction and prompting strategies affect video quality across tasks.

\subsection{Evaluation Setup}
\textbf{Analyzed Components and Baselines.}
We analyze three design factors and combine their labels to denote experimental configurations (e.g., \texttt{G1/P1/PE2}). Implementation details are provided in Appendix~\ref{sec:empirical_implementation}.
\begin{enumerate}
    \item \textbf{Proxy granularity (\texttt{G}):} \texttt{G0}, \texttt{G1}, and \texttt{G2} denote coarse, medium, and fine geometry and motion, respectively.

    \item \textbf{Physics requirements (\texttt{P}):} \texttt{P1} requires Blender's built-in physics simulation; \texttt{P0} leaves its use optional.

    \item \textbf{Prompt enhancement and proxy alignment (\texttt{PE}):} \texttt{PE0} applies no additional enhancement; \texttt{PE1} uses MiniMax H3-Context-IR \citep{minimax2026h3contextir} to enrich the prompt; and \texttt{PE2} requires the coding agent itself to specify subject--color bindings in the prompt and add complementary appearance details based on the constructed proxy.
\end{enumerate}
We use Seedance 2.5 as the direct generation baseline, conditioned on the original prompt and optional references.

\begin{figure*}[p]
    \centering
    \includegraphics[
        width=\textwidth,
        height=0.85\textheight,
        keepaspectratio
    ]{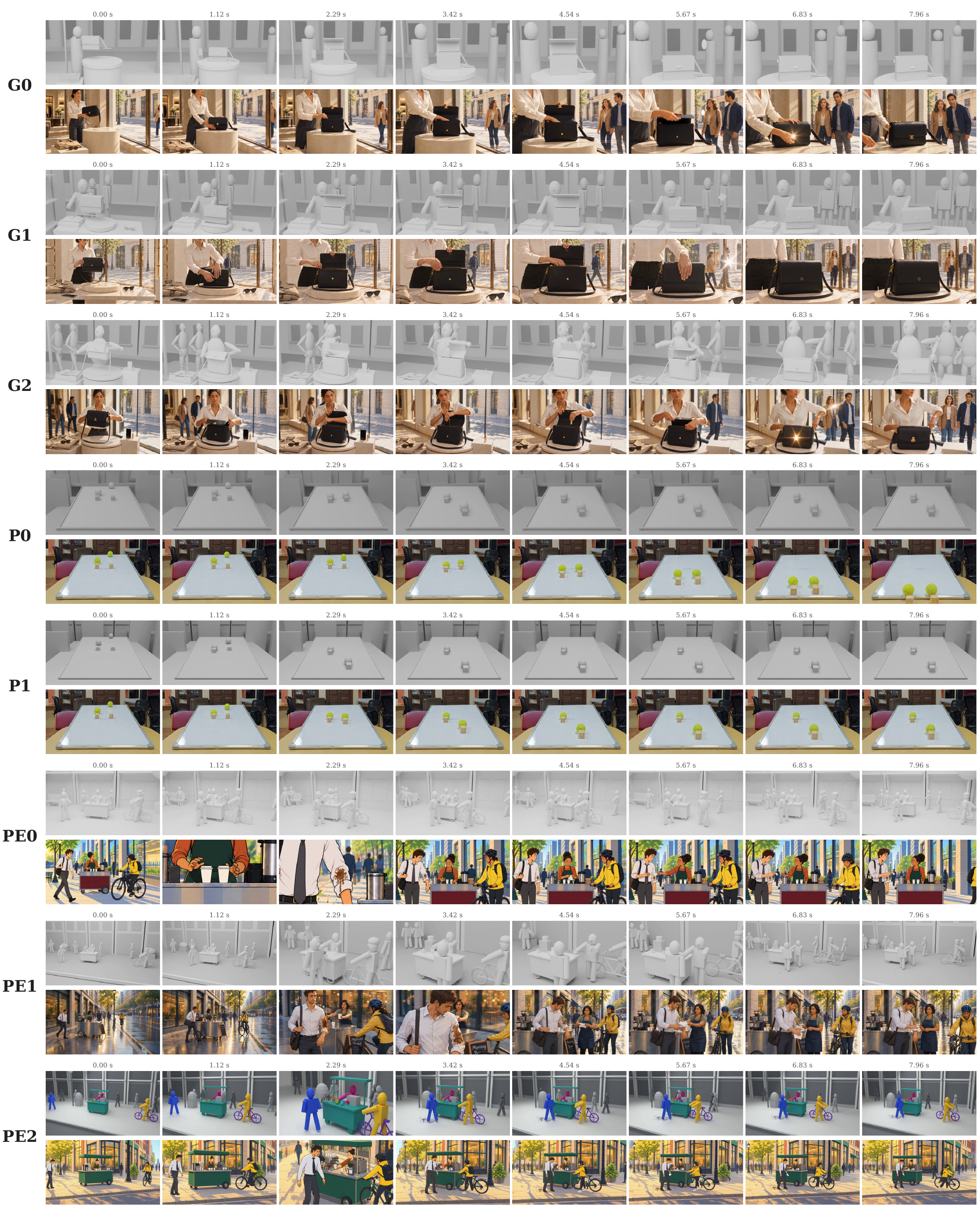}
    \caption{\textbf{Qualitative visualization of component configurations.} Within each group, the upper sequence shows the proxy and the lower sequence shows the generated video at the same eight timestamps. The physics examples labeled \texttt{P0} and \texttt{P1} in the image actually correspond to \texttt{G1/P1/PE1} and \texttt{G0/P1/PE0}, respectively. Both require physics simulation and differ in granularity and prompting; they are illustrative examples, not a controlled \texttt{P0}/\texttt{P1} comparison.}
    \label{fig:component_visualization}
\end{figure*}

\textbf{Visualization of component implementations.}
Figure~\ref{fig:component_visualization} compares proxy videos with their generated outputs. The \texttt{G0}--\texttt{G2} rows illustrate changes in geometry and articulation, while the \texttt{PE0}--\texttt{PE2} rows illustrate different prompting strategies. The two physics examples use different proxy configurations and do not isolate the physics requirement. For each configuration, we uniformly sample eight timestamps over the shared video interval and align proxy and generated frames at identical timestamps to inspect the transfer of placement, motion, interactions, and appearance.

\textbf{Tasks.}
Our evaluation covers three task categories:
\begin{enumerate}
    \item \textbf{WorldMemBench:} Scene revisitation, evaluating consistency on return visits.

    \item \textbf{VGIF-Bench:} Complex instruction following involving human or object motion and interactions.

    \item \textbf{Principia:} Physical process simulation, which evaluates videos depicting physical phenomena.
\end{enumerate}

\textbf{Evaluation metrics and protocol.}
Experts score each video on four dimensions under a blinded protocol using $\{0,1,2,3\}$ (higher is better).
\begin{enumerate}
    \item \textbf{Consistency:} Temporal consistency within the generated video and consistency with the input reference.

    \item \textbf{Instruction following:} The extent to which the generated video satisfies the input instruction.

    \item \textbf{Physical plausibility:} The physical plausibility of the depicted motion, interactions, and processes.

    \item \textbf{Naturalness and visual realism:} The perceived naturalness of motion and overall visual realism.
\end{enumerate}
The implementation details of the empirical analysis are in Appendix~\ref{sec:empirical_implementation}.

\textbf{Samples and aggregation.}
The evaluation set comprises 94 cases from WorldMemBench (33), VGIF-Bench (29), and Principia (32). In Table~\ref{tab:human_evaluation_ablation}, we report per-metric means over the available ratings for each configuration, pooling cases across benchmarks for the Average columns. Execution environment errors reduced the coverage of \texttt{G1/P1/PE1} to 80 cases, including 18 from Principia. For \texttt{G0/P0/PE0}, one missing VGIF-Bench consistency rating yields 93 valid ratings for this metric; the other metrics cover all 94 cases. Four-metric means are computed over cases with complete ratings.

\textbf{Paired comparisons and scope.}
Tables~\ref{tab:physics_ablation_deltas} and~\ref{tab:pe_ablation_deltas} use each configuration pair's common cases with complete ratings on all four metrics. For each metric, we average per-case score differences over this shared subset, equivalently subtracting the two means on the same subset. Mean averages these four differences. Calculations precede rounding, so the deltas need not equal subtraction of Table~\ref{tab:human_evaluation_ablation}'s displayed values. These results describe quality on common successful cases, not performance including generation failures. Raters, run batches, and retry budgets are not fully matched, so comparisons are descriptive. \texttt{PE2} evaluates prompt construction and subject--color binding jointly.

\subsection{Evaluation Results}

\begin{figure}[htbp]
    \centering
    \includegraphics[width=\linewidth]{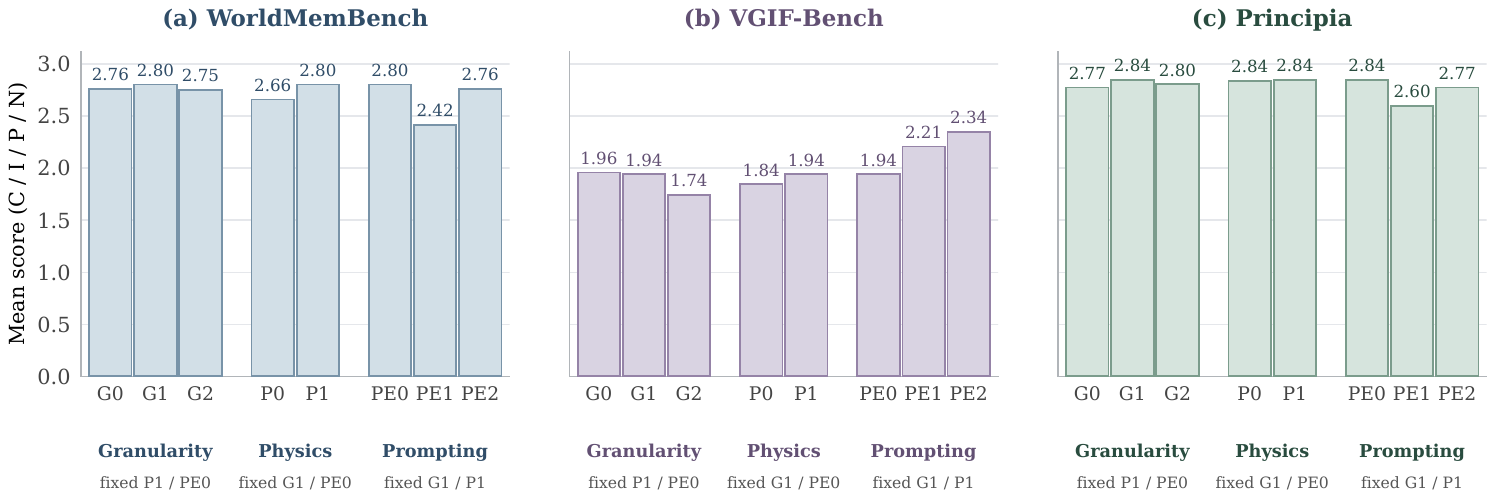}
    \caption{\textbf{Effects of proxy granularity, physics simulation, and prompt enhancement on three benchmarks.} Each bar averages the four scores (C/I/P/N) for one configuration and benchmark in Table~\ref{tab:human_evaluation_ablation}. From left to right, the groups vary granularity at fixed \texttt{P1/PE0}, physics requirements at fixed \texttt{G1/PE0}, and prompting at fixed \texttt{G1/P1}, sharing \texttt{G1/P1/PE0} as a reference. These are available-case summaries: \texttt{G1/P1/PE1} uses 18 Principia cases, whereas the other configurations shown use 32. Matched comparisons are reported separately in Tables~\ref{tab:physics_ablation_deltas} and~\ref{tab:pe_ablation_deltas}.}
    \label{fig:benchmark_component_means}
\end{figure}

\begin{table*}[t]
\centering
\caption{Human evaluation using all available ratings on three benchmarks and per-metric averages over their pooled cases. C/I/P/N denote consistency, instruction following, physical plausibility, and naturalness and visual realism. Sample sizes vary as specified in the evaluation protocol; \texttt{G1/P1/PE1} covers only 80 of 94 cases. Bold marks column maxima. Blue/red shading indicates scores above/at or below direct generation; darker shading denotes larger differences.}
\label{tab:human_evaluation_ablation}

\begingroup
\color{black}
\footnotesize
\setlength{\tabcolsep}{2.4pt}
\renewcommand{\arraystretch}{1.20}

\resizebox{0.95\textwidth}{!}{%
\begin{tabular}{@{}lcccc@{\hspace{8pt}}cccc@{\hspace{8pt}}cccc@{\hspace{8pt}}cccc}
\toprule
& \multicolumn{4}{c}{\textbf{WorldMemBench}}
& \multicolumn{4}{c}{\textbf{VGIF-Bench}}
& \multicolumn{4}{c}{\textbf{Principia}}
& \multicolumn{4}{c}{\textbf{Average}} \\
\cmidrule(lr){2-5}
\cmidrule(lr){6-9}
\cmidrule(lr){10-13}
\cmidrule(lr){14-17}
\textbf{Configuration}
& C$\uparrow$ & I$\uparrow$ & P$\uparrow$ & N$\uparrow$
& C$\uparrow$ & I$\uparrow$ & P$\uparrow$ & N$\uparrow$
& C$\uparrow$ & I$\uparrow$ & P$\uparrow$ & N$\uparrow$
& C$\uparrow$ & I$\uparrow$ & P$\uparrow$ & N$\uparrow$ \\
\midrule

Direct generation
& 2.182 & 1.242 & 2.485 & 2.333
& 2.483 & 1.828 & \textbf{2.207} & \textbf{2.000}
& 2.562 & 2.406 & 2.000 & 2.156
& 2.404 & 1.819 & 2.234 & 2.170 \\

\midrule

\texttt{G0/P0/PE0}
& \HEgain{23.69}{2.758} & \HEgain{60.00}{\textbf{3.000}} & \HEgain{9.72}{2.606} & \HEgain{9.75}{2.455}
& \HEnongain{7.66}{2.429} & \HEgain{19.76}{2.276} & \HEnongain{15.52}{1.897} & \HEnongain{24.00}{1.414}
& \HEgain{18.50}{2.969} & \HEgain{22.34}{2.938} & \HEgain{23.26}{2.562} & \HEgain{19.45}{2.594}
& \HEgain{16.04}{2.731} & \HEgain{34.75}{2.755} & \HEgain{10.24}{2.372} & \HEgain{6.34}{2.181} \\

\texttt{G1/P0/PE0}
& \HEgain{19.02}{2.606} & \HEgain{60.00}{\textbf{3.000}} & \HEgain{8.80}{2.576} & \HEgain{9.75}{2.455}
& \HEgain{12.36}{2.690} & \HEgain{17.64}{2.207} & \HEnongain{29.31}{1.448} & \HEnongain{35.67}{1.034}
& \HEgain{18.50}{2.969} & \HEgain{24.25}{\textbf{3.000}} & \HEgain{26.15}{2.656} & \HEgain{23.29}{2.719}
& \HEgain{16.78}{2.755} & \HEgain{34.75}{2.755} & \HEgain{6.65}{2.255} & \HEnongain{7.97}{2.106} \\

\texttt{G2/P0/PE0}
& \HEgain{24.61}{2.788} & \HEgain{60.00}{\textbf{3.000}} & \HEgain{9.72}{2.606} & \HEgain{9.75}{2.455}
& \HEgain{8.12}{2.552} & \HEgain{13.40}{2.069} & \HEnongain{29.31}{1.448} & \HEnongain{36.72}{1.000}
& \HEgain{16.57}{2.906} & \HEgain{20.41}{2.875} & \HEgain{22.31}{2.531} & \HEgain{18.47}{2.562}
& \HEgain{16.78}{2.755} & \HEgain{32.14}{2.670} & \HEnongain{6.34}{2.223} & \HEnongain{9.90}{2.043} \\

\texttt{G0/P1/PE0}
& \HEgain{19.02}{2.606} & \HEgain{59.08}{2.970} & \HEgain{14.39}{2.758} & \HEgain{17.18}{\textbf{2.697}}
& \HEnongain{7.08}{2.448} & \HEgain{19.76}{2.276} & \HEnongain{19.76}{1.759} & \HEnongain{26.12}{1.345}
& \HEgain{18.50}{2.969} & \HEgain{23.29}{2.969} & \HEgain{22.31}{2.531} & \HEgain{20.41}{2.625}
& \HEgain{14.51}{2.681} & \HEgain{34.75}{2.755} & \HEgain{10.24}{2.372} & \HEgain{8.61}{\textbf{2.255}} \\

\texttt{G1/P1/PE0}
& \HEgain{25.54}{\textbf{2.818}} & \HEgain{60.00}{\textbf{3.000}} & \HEgain{15.31}{\textbf{2.788}} & \HEgain{14.39}{2.606}
& \HEgain{9.16}{2.586} & \HEgain{19.76}{2.276} & \HEnongain{21.88}{1.690} & \HEnongain{30.36}{1.207}
& \HEgain{18.50}{2.969} & \HEgain{22.34}{2.938} & \HEgain{27.13}{2.688} & \HEgain{25.20}{\textbf{2.781}}
& \HEgain{18.10}{2.798} & \HEgain{34.75}{2.755} & \HEgain{11.56}{2.415} & \HEgain{7.97}{2.234} \\

\texttt{G2/P1/PE0}
& \HEgain{22.74}{2.727} & \HEgain{60.00}{\textbf{3.000}} & \HEgain{13.43}{2.727} & \HEgain{12.51}{2.545}
& \HEnongain{8.12}{2.414} & \HEgain{18.69}{2.241} & \HEnongain{31.43}{1.379} & \HEnongain{38.84}{0.931}
& \HEgain{18.50}{2.969} & \HEgain{22.34}{2.938} & \HEgain{29.04}{\textbf{2.750}} & \HEgain{18.47}{2.562}
& \HEgain{15.49}{2.713} & \HEgain{34.44}{2.745} & \HEgain{8.61}{2.319} & \HEnongain{9.59}{2.053} \\

\texttt{G1/P0/PE1}
& \HEgain{14.39}{2.455} & \HEgain{59.08}{2.970} & \HEgain{10.64}{2.636} & \HEgain{14.39}{2.606}
& \HEnongain{8.12}{2.414} & \HEgain{16.57}{2.172} & \HEnongain{32.48}{1.345} & \HEnongain{39.88}{0.897}
& \HEnongain{21.36}{2.062} & \HEgain{7.94}{2.469} & \HEgain{19.45}{2.438} & \HEgain{9.84}{2.281}
& \HEnongain{8.92}{2.309} & \HEgain{28.55}{2.553} & \HEnongain{7.97}{2.170} & \HEnongain{12.20}{1.968} \\

\texttt{G1/P0/PE2}
& \HEgain{24.61}{2.788} & \HEgain{60.00}{\textbf{3.000}} & \HEgain{8.80}{2.576} & \HEgain{12.51}{2.545}
& \HEgain{9.16}{2.586} & \HEgain{20.81}{2.310} & \HEnongain{29.31}{1.448} & \HEnongain{35.67}{1.034}
& \HEgain{17.55}{2.938} & \HEgain{23.29}{2.969} & \HEgain{22.31}{2.531} & \HEgain{21.36}{2.656}
& \HEgain{17.46}{2.777} & \HEgain{35.43}{2.777} & \HEnongain{6.65}{2.213} & \HEnongain{7.63}{2.117} \\

\texttt{G1/P1/PE1}
& \HEnongain{6.00}{2.182} & \HEgain{50.69}{2.697} & \HEgain{6.92}{2.515} & \HEnongain{7.84}{2.273}
& \HEgain{13.40}{2.724} & \HEgain{38.84}{2.897} & \HEnongain{17.64}{1.828} & \HEnongain{25.08}{1.379}
& \HEgain{19.45}{\textbf{3.000}} & \HEgain{24.25}{\textbf{3.000}} & \HEgain{9.41}{2.111} & \HEgain{9.75}{2.278}
& \HEgain{10.85}{2.562} & \HEgain{37.27}{2.837} & \HEnongain{7.81}{2.175} & \HEnongain{12.76}{1.950} \\

\texttt{G1/P1/PE2}
& \HEgain{25.54}{\textbf{2.818}} & \HEgain{59.08}{2.970} & \HEgain{13.43}{2.727} & \HEgain{11.59}{2.515}
& \HEgain{20.84}{\textbf{2.966}} & \HEgain{39.88}{\textbf{2.931}} & \HEnongain{11.31}{2.034} & \HEnongain{22.96}{1.448}
& \HEgain{19.45}{\textbf{3.000}} & \HEgain{24.25}{\textbf{3.000}} & \HEgain{21.36}{2.500} & \HEgain{19.45}{2.594}
& \HEgain{22.03}{\textbf{2.926}} & \HEgain{41.29}{\textbf{2.968}} & \HEgain{12.20}{\textbf{2.436}} & \HEgain{7.32}{2.213} \\

\bottomrule
\end{tabular}%
}

\endgroup
\end{table*}

Table~\ref{tab:human_evaluation_ablation} summarizes the results across configurations and benchmarks. We focus on trends that recur across settings and evaluation dimensions. Figure~\ref{fig:benchmark_component_means} visualizes available-case trends within each benchmark; paired comparisons below account for differences in case coverage.

\begin{observationbox}[Observation 1: Favorable and Challenging Scenarios of Code2Video]
Code2Video most consistently improves \textbf{consistency} and \textbf{instruction following} over direct generation. Its weaknesses concern \textbf{physical plausibility}, \textbf{motion fluency}, and \textbf{realism} on \textbf{VGIF-Bench}, with extensive human and object motion.
\end{observationbox}
Under \texttt{PE0}, all six configurations outperform Seedance on all four dimensions on WorldMemBench and Principia (Table~\ref{tab:human_evaluation_ablation}). On VGIF-Bench, all improve instruction following but score lower in physical plausibility, naturalness, and realism, with mixed consistency results. All six also outperform Seedance in consistency and instruction following averaged over samples from the three benchmarks.

\begin{observationbox}[Observation 2: Granularity Trade-offs]
With \texttt{PE0} fixed, higher modeling and motion detail does not necessarily improve video quality. Coarse proxies (\texttt{G0}) favor \textbf{naturalness and physical plausibility} on VGIF, while medium-grained proxies (\texttt{G1}) achieve the highest or tied-highest \textbf{consistency} across all three benchmarks under \texttt{P1}.
\end{observationbox}
Table~\ref{tab:human_evaluation_ablation} shows that, with \texttt{PE0} fixed, \texttt{G2} never exceeds \texttt{G1} in naturalness and realism across benchmarks and physics settings. On VGIF-Bench, \texttt{G0} leads the three granularities in physical plausibility and naturalness under both physics settings. Manual inspection attributes this advantage primarily to leaving limb motion and fine details to the video generator, reducing stiffness inherited from detailed proxy animation. Under \texttt{P1}, \texttt{G1} improves consistency over \texttt{G0} on WorldMemBench and VGIF while tying on Principia.

\begin{observationbox}[Observation 3: Benefits of Physics Simulation]
On each pair's common complete-case subset, explicit physics requirements (\texttt{P1}) improve the \textbf{four-metric mean over cases pooled from the three benchmarks} across all five tested comparisons. Gains are more consistent in \textbf{physical plausibility} and \textbf{naturalness and realism} than in \textbf{consistency}.
\end{observationbox}

\begin{table}[htbp]
\centering
\caption{Paired \texttt{P1}-minus-\texttt{P0} score changes on common cases with complete ratings, pooled across three benchmarks. Sample sizes are 93 for \texttt{G0/PE0}, 80 for \texttt{G1/PE1}, and 94 otherwise. Mean averages C/I/P/N changes. Differences are computed before rounding.}
\label{tab:physics_ablation_deltas}
\begingroup
\color{black}
\footnotesize
\setlength{\tabcolsep}{2.5pt}
\renewcommand{\arraystretch}{1.15}
\resizebox{0.5\columnwidth}{!}{%
\begin{tabular}{@{}llrrrrr}
\toprule
\textbf{Comparison} & \textbf{Config.} & \textbf{C} & \textbf{I} & \textbf{P} & \textbf{N} & \textbf{Mean} \\
\midrule
\multirow{5}{*}{\texttt{P1-P0}} & \texttt{G0/PE0} & \HEnongain{17.66}{-0.054} & 0.000 & 0.000 & \HEgain{22.20}{+0.075} & \HEgain{7.08}{+0.005} \\
& \texttt{G1/PE0} & \HEgain{15.29}{+0.043} & 0.000 & \HEgain{40.56}{+0.160} & \HEgain{33.65}{+0.128} & \HEgain{23.71}{+0.082} \\
& \texttt{G2/PE0} & \HEnongain{15.29}{-0.043} & \HEgain{21.98}{+0.074} & \HEgain{26.74}{+0.096} & \HEgain{8.38}{+0.011} & \HEgain{13.56}{+0.035} \\
& \texttt{G1/PE1} & \HEgain{49.20}{+0.200} & \HEgain{60.00}{+0.250} & \HEgain{19.39}{+0.062} & \HEgain{8.59}{+0.012} & \HEgain{34.30}{+0.131} \\
& \texttt{G1/PE2} & \HEgain{38.18}{+0.149} & \HEgain{47.26}{+0.191} & \HEgain{54.17}{+0.223} & \HEgain{26.74}{+0.096} & \HEgain{41.64}{+0.165} \\
\bottomrule
\end{tabular}%
}
\endgroup
\end{table}

Table~\ref{tab:physics_ablation_deltas} shows higher four-metric means and naturalness and realism scores in all five comparisons on their respective common complete-case subsets. Physical plausibility and instruction following improve or remain unchanged, whereas consistency changes in both directions. \texttt{G1/PE2} has the largest reported gains in the four-metric mean and physical plausibility; the differing sample subsets preclude interpreting this ranking as a controlled interaction effect.

\begin{observationbox}[Observation 4: Benefits of Agent-guided PE]
With \texttt{G1} fixed, paired comparisons on common complete cases show higher pooled scores for \texttt{PE2} than \texttt{PE1} in all four metrics under both physics settings. Under \texttt{P1}, \texttt{PE2} also improves the four-metric mean over \texttt{PE0}, with gains in \textbf{consistency}, \textbf{instruction following}, and \textbf{physical plausibility}.
\end{observationbox}

\begin{table}[htbp]
\centering
\caption{Paired score changes from \texttt{PE2} at fixed \texttt{G1}, pooled across three benchmarks. Each row uses that pair's common complete cases: 80 for \texttt{PE2-PE1} under \texttt{P1}, and 94 for the other rows. Mean averages C/I/P/N changes. Differences are computed before rounding.}
\label{tab:pe_ablation_deltas}
\begingroup
\color{black}
\footnotesize
\setlength{\tabcolsep}{2.5pt}
\renewcommand{\arraystretch}{1.15}
\resizebox{0.5\columnwidth}{!}{%
\begin{tabular}{@{}llrrrrr}
\toprule
\textbf{Comparison} & \textbf{Config.} & \textbf{C} & \textbf{I} & \textbf{P} & \textbf{N} & \textbf{Mean} \\
\midrule
\multirow{2}{*}{\texttt{PE2-PE1}}
& \texttt{G1/P0} & \HEgain{60.00}{+0.468} & \HEgain{31.73}{+0.223} & \HEgain{10.96}{+0.043} & \HEgain{23.19}{+0.149} & \HEgain{31.50}{+0.221} \\
& \texttt{G1/P1} & \HEgain{46.38}{+0.350} & \HEgain{20.42}{+0.125} & \HEgain{33.46}{+0.238} & \HEgain{24.69}{+0.162} & \HEgain{31.27}{+0.219} \\
\midrule
\multirow{2}{*}{\texttt{PE2-PE0}}
& \texttt{G1/P0} & \HEgain{8.42}{+0.021} & \HEgain{8.42}{+0.021} & \HEnongain{10.96}{-0.043} & \HEgain{7.27}{+0.011} & \HEgain{6.35}{+0.003} \\
& \texttt{G1/P1} & \HEgain{20.77}{+0.128} & \HEgain{30.58}{+0.213} & \HEgain{8.42}{+0.021} & \HEnongain{8.42}{-0.021} & \HEgain{15.81}{+0.085} \\
\bottomrule
\end{tabular}%
}
\endgroup
\end{table}

\begin{figure}[htbp]
    \centering
    \includegraphics[width=\linewidth]{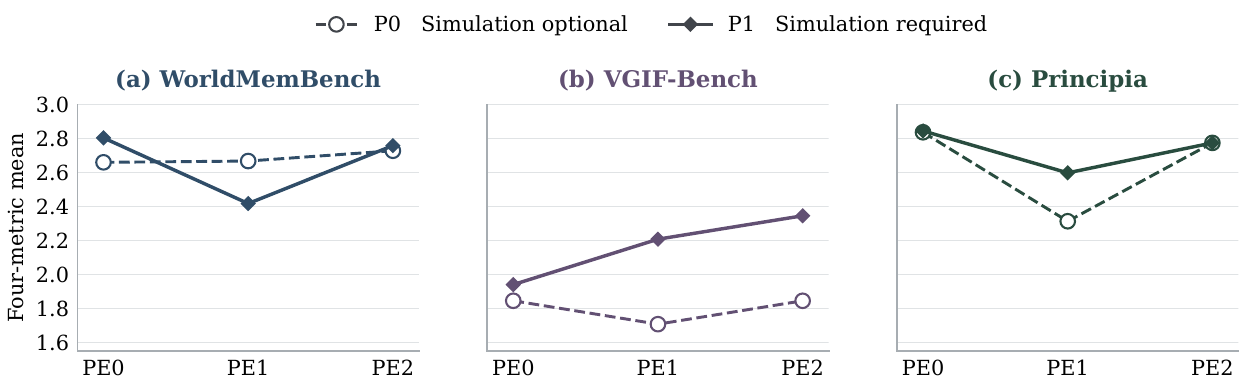}
    \caption{\textbf{Prompt enhancement under two physics settings at fixed \texttt{G1}.} Each point averages the four scores (C/I/P/N) for one benchmark and configuration in Table~\ref{tab:human_evaluation_ablation}. Dashed lines with open circles denote optional physics simulation (\texttt{P0}); solid lines with diamonds denote required simulation (\texttt{P1}). These are available-case summaries, not uniformly matched comparisons: \texttt{G1/P1/PE1} uses 18 Principia cases, while the other configurations use 32.}
    \label{fig:physics_prompt_interaction}
\end{figure}

Table~\ref{tab:pe_ablation_deltas} confirms that \texttt{PE2} outperforms \texttt{PE1} in all four metrics on their shared cases under both physics settings. The \texttt{PE2-PE1} comparison under \texttt{P1} uses 80 cases; the other three comparisons use 94. Relative to \texttt{PE0}, \texttt{PE2} yields a larger four-metric gain under \texttt{P1}, improving consistency, instruction following, and physical plausibility despite a slight decline in naturalness and realism. Figure~\ref{fig:physics_prompt_interaction} shows how the effect of prompt enhancement varies across benchmarks at fixed \texttt{G1}. On VGIF-Bench, switching from \texttt{PE0} to \texttt{PE2} raises the four-metric mean by 0.405 points under \texttt{P1}, while the mean is unchanged under \texttt{P0}. On WorldMemBench and Principia, the same switch slightly lowers the mean under \texttt{P1}.

\begin{observationbox}[Observation 5: Best Reported Overall Configuration]
In the available-case summaries, \texttt{G1/P1/PE2} achieves the \textbf{highest reported four-metric mean over cases pooled from the three benchmarks}, together with the highest corresponding consistency, instruction-following, and physical-plausibility averages.
\end{observationbox}

\begin{figure*}[t]
    \centering
    \includegraphics[width=\textwidth]{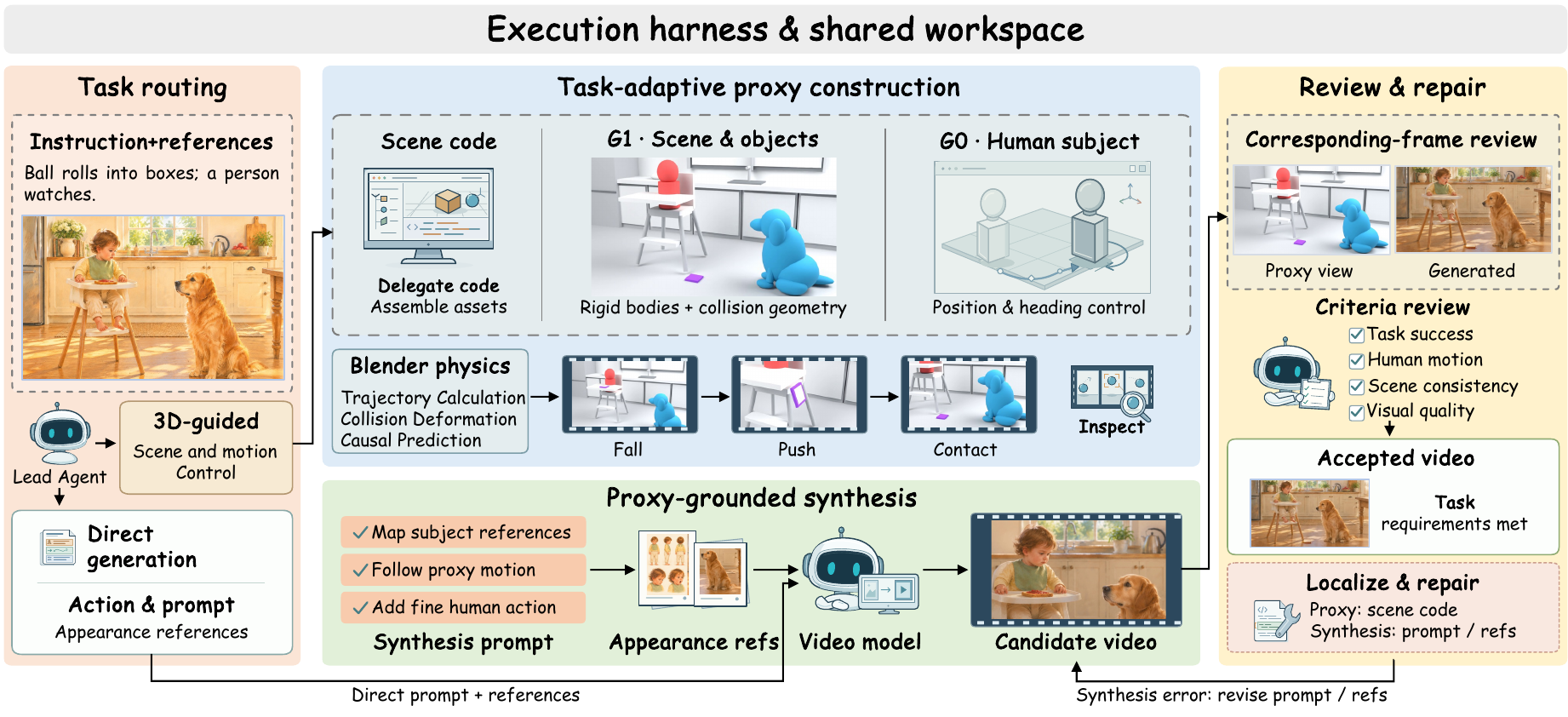}
    \caption{\textbf{Overview of \name.} \name combines task-adaptive proxy construction with proxy-grounded video synthesis. The harness maintains the working scene and supports visual feedback throughout generation.}
    \label{fig:method}
\end{figure*}

The available-case scores in Table~\ref{tab:human_evaluation_ablation} place \texttt{G1/P1/PE2} first in the four-metric mean and in consistency, instruction following, and physical plausibility. Its naturalness and realism score exceeds Seedance but trails \texttt{G0/P1/PE0}. This is a descriptive ranking under unequal case coverage; the paired comparisons in Tables~\ref{tab:physics_ablation_deltas} and~\ref{tab:pe_ablation_deltas} provide the corresponding component-level evidence.

\subsection{Empirical Answers and Workflow Insights}
The above empirical observations provide the following answers to the two central questions.

\begin{insightbox}[Answers to the Central Questions]
\begin{enumerate}
    \item \textbf{How far can Code2Video extend video generation?} Code2Video most consistently improves consistency and instruction following over direct generation, with broader gains on scene revisitation and physical-process simulation. Its benefits remain task-dependent: complex human and object motion on VGIF-Bench still presents weaknesses in physical plausibility, naturalness, and realism.

    \item \textbf{How should the two models' strengths be combined?} Coding agents should specify essential scene structure and dynamics, while video-generation models realize appearance and fine motion. The configuration with the highest available-case mean, \texttt{G1/P1/PE2}, combines medium-grained proxies, required physics simulation, and agent-guided prompt enhancement. The advantages of coarse proxies on motion-intensive tasks further motivate adapting the allocation of control to the task.
\end{enumerate}
\end{insightbox}

\textbf{Workflow Design Insights.}
These findings motivate three coordinated workflow choices: adapt modeling and motion-control granularity to each entity, favoring coarse proxies for human motion and medium-grained proxies where structural consistency matters; selectively invoke Blender's built-in physics engine for interactions the proxy can meaningfully simulate; and coordinate proxy construction with prompt enhancement by establishing subject--color bindings during modeling and explicitly describing these bindings alongside the proxy's geometry, motion, and simulated behavior in the generation prompt.
\section{Method}
\label{sec:method}
\label{sec:method_overview}

We introduce \textbf{\name}, a task-adaptive agentic framework for code2video generation. 
\name selects a generation path based on task requirements and applies the corresponding modeling and prompting skills.
An execution harness coordinates scene construction and video synthesis, using visual feedback to guide revisions, as shown in Figure~\ref{fig:method}.

\subsection{Task-Adaptive Proxy Construction}
\label{sec:method_routing}

\textbf{Route selection.}
Based on the input instruction and reference media, \name selects 3D modeling for tasks requiring precise control over scene structure or motion. 
For tasks centered on natural body movement, it selects direct video generation when explicit control is unnecessary and simplified proxy animation could introduce stiffness. 
Both routes support video synthesis, with Blender tools additionally enabled for 3D modeling.

\textbf{Modeling granularity.}
We encode the entity-wise policy from the preceding analysis in modeling skills: \texttt{G1} for environments and structured objects, and \texttt{G0} for human subjects. 
The scene code specifies overall human movement, while the synthesis prompt describes finer actions. 
The agent simulates requested physical interactions when they can be meaningfully represented in Blender.

\subsection{Proxy-Grounded Video Synthesis}
\label{sec:method_pe}

\textbf{From proxy to synthesis prompt.}
After constructing the proxy, the agent inspects the rendered sequence and revises the scene code when the depicted events deviate from the request. 
Guided by the synthesis skill, it then uses the inspected proxy to write a prompt describing the intended final video. 
The prompt links subject descriptions to the corresponding proxy entities and appearance references, follows the actions depicted in the proxy, and specifies the visual details and subtle movements to be synthesized.

\textbf{Video synthesis.}
Before synthesis, the harness checks that the proxy video and its shot description correspond to the same render and resolves subject references. 
The generation tool then passes the prompt, proxy video, and associated reference media to the video model, which produces a candidate video. 
The direct-generation path uses the same synthesis stage without a proxy, supplying a task-derived action and camera prompt with the available references.

\subsection{Harness Design and Visual Feedback}
\label{sec:method_acceptance}

\textbf{Workflow execution.}
The harness provides a shared Blender workspace and tool interfaces. 
A lead generation agent coordinates execution and can delegate separable modeling tasks to coding subagents. 
Each assignment specifies how a component should fit into the scene, and returned assets are then integrated and inspected in context. 
The workspace preserves this work across revisions, allowing subagents to continue refining their components and supporting asset reuse in later video extensions.

\textbf{Visual feedback.}
Visual inspection covers both the proxy and the generated video. 
For code2video outputs, the harness displays generated frames alongside the corresponding proxy views to help identify the source of discrepancies. 
Proxy errors lead to scene revisions, while discrepancies introduced during synthesis guide changes to the prompt or reference inputs. 
An independent acceptance agent reviews each candidate against the original task and its registered criteria.
The harness links the review to the submitted video and returns unmet requirements for repair.

\begin{table*}[t]
\centering
\caption{\textbf{Main results across four benchmarks.} R2M reports normalized memory ratios (NMR). WER: word error rate; VR: video realism; IF: visual instruction following. Green cells mark the best result in each column.}
\label{tab:main_results}
\footnotesize
\setlength{\tabcolsep}{3pt}
\renewcommand{\arraystretch}{1.08}
\resizebox{\textwidth}{!}{%
\begin{tabular}{@{}lccccccccc}
\toprule
& \textbf{Last Exam} & \multicolumn{4}{c}{\textbf{R2M}} & \multicolumn{2}{c}{\textbf{VTR}} & \multicolumn{2}{c}{\textbf{T2AV-Compass}} \\
\cmidrule(lr){3-6} \cmidrule(lr){7-8} \cmidrule(lr){9-10}
\textbf{Method} & Score$\uparrow$ & Scene ID$\uparrow$ & Local Geo.$\uparrow$ & State$\uparrow$ & Object$\uparrow$ & Checklist$\uparrow$ & WER$\downarrow$ & VR$\uparrow$ & IF$\uparrow$ \\
\midrule
Direct generation & 47.16 & 0.347 & 0.351 & 0.159 & 0.343 & 85.15 & 59.21 & 71.38 & 65.11 \\
\midrule
\rowcolor{groupblue}
\multicolumn{10}{c}{\rule{0pt}{2.3ex}\textit{\textbf{Agentic Video Generation}}} \\
\hspace{0.6em}Codex & 53.18 & 0.575 & 0.583 & 0.520 & 0.587 & 89.72 & 25.73 & \cellcolor{bestgreen}72.21 & 68.05 \\
\hspace{0.6em}Claude & 51.95 & 0.594 & 0.590 & 0.523 & 0.522 & 87.07 & 53.62 & 71.50 & \cellcolor{bestgreen}68.22 \\
\hspace{0.6em}Codex + VideoWeaver & 55.11 & 0.739 & 0.705 & 0.646 & 0.665 & 90.88 & 23.58 & 71.83 & 67.60 \\
\hspace{0.6em}Claude + VideoWeaver & 48.45 & 0.658 & 0.652 & 0.659 & 0.560 & 89.87 & 55.06 & 71.54 & 68.19 \\
\midrule
\rowcolor{groupblue}
\multicolumn{10}{c}{\rule{0pt}{2.3ex}\textit{\textbf{Code2Video}}} \\
\hspace{0.6em}Codex + Blender & 59.49 & 0.936 & 0.840 & 0.743 & 0.860 & 88.52 & 22.30 & 67.29 & 64.51 \\
\hspace{0.6em}Claude + Blender & \cellcolor{bestgreen}65.52 & 0.950 & 0.847 & 0.823 & 0.805 & 88.23 & 50.88 & 66.54 & 64.28 \\
\hspace{0.6em}VideoCoCo & 59.37 & 0.931 & 0.853 & 0.696 & 0.829 & 83.67 & 27.11 & 66.28 & 61.61 \\
\hspace{0.6em}CodeVideoModel & 53.86 & 0.932 & 0.878 & 0.708 & 0.953 & 84.75 & 20.74 & 64.88 & 61.50 \\
\hspace{0.6em}\textbf{\name (Ours)} & 60.77 & \cellcolor{bestgreen}1.011 & \cellcolor{bestgreen}0.947 & \cellcolor{bestgreen}0.875 & \cellcolor{bestgreen}0.966 & \cellcolor{bestgreen}91.42 & \cellcolor{bestgreen}11.08 & 70.26 & 67.99 \\
\bottomrule
\end{tabular}}
\end{table*}

\section{Experiments}
\label{sec:experiment}

\subsection{Experimental Setup}
We evaluate \name on World Models' Last Exam in Physics~\cite{gao2026lastexam}, R2M~\cite{gu2026r2m}, and VTR~\cite{huang2026vtr} for physical behavior, visual consistency, and detailed instruction following, respectively.
We further assess video realism and visual instruction following on 200 randomly sampled T2AV-Compass tasks~\cite{cao2026t2avcompass}.
All methods use Seedance 2.5~\cite{seedance2026seedance} for video synthesis, with benchmark images supplied through the reference-media interface to ensure consistent input.

\textbf{Baselines.}
We compare three groups: (1) direct generation using task instructions and reference media; (2) agentic generation with Codex or Claude equipped with video-generation tools and skills, with or without VideoWeaver~\cite{wei2026videoweaver}; and (3) Code2Video methods, including Codex + Blender, Claude + Blender, VideoCoCo~\cite{li2026videococo}, CodeVideoModel~\cite{liang2026codevideomodel}, and \name.
Code2Video methods render proxy videos to guide Seedance synthesis.

\textbf{Models.}
All Codex-based methods, VideoCoCo, CodeVideoModel, and \name use GPT-6-Astra~\cite{openai2026gpt6astra}, while Claude-based methods use Claude Opus 5.5~\cite{anthropic2026claudeopus55}.
\name and CodeVideoModel additionally uses GPT-Image-2 for reference-image generation.

\begin{table}[t]
\centering
\caption{\textbf{Ablations on the test subset.} P: physics; I: instruction following; C: consistency; A\&F: aesthetics and fluency. Green marks column maxima (including ties) per video model.}
\label{tab:agent_model_ablation}
\small
\setlength{\tabcolsep}{3pt}
\renewcommand{\arraystretch}{1.12}
\begin{tabular*}{\columnwidth}{@{\extracolsep{\fill}}lccccc@{}}
\toprule
\textbf{Agent} & \textbf{P} & \textbf{I} & \textbf{C} & \textbf{A\&F} & \textbf{Mean} \\
\midrule
\rowcolor{groupblue}
\multicolumn{6}{c}{\textit{\textbf{Seedance 2.5}}} \\
Direct generation & 3.242 & 3.364 & 3.758 & 3.667 & 3.508 \\
GPT-6-Astra & 3.455 & \cellcolor{bestgreen}3.758 & \cellcolor{bestgreen}3.879 & \cellcolor{bestgreen}3.758 & \cellcolor{bestgreen}3.712 \\
GPT-6.1-Sol & 3.394 & 3.545 & 3.818 & 3.667 & 3.606 \\
Claude Opus 5.5 & 3.548 & 3.742 & 3.839 & 3.645 & 3.694 \\
Gemini 4 Argon & \cellcolor{bestgreen}3.562 & 3.719 & 3.656 & 3.656 & 3.648 \\
\midrule
\rowcolor{groupblue}
\multicolumn{6}{c}{\textit{\textbf{MiniMax H3}}} \\
Direct generation & 3.333 & 3.545 & 3.636 & 3.606 & 3.530 \\
GPT-6-Astra & 3.394 & 3.485 & 3.697 & 3.667 & 3.561 \\
GPT-6.1-Sol & \cellcolor{bestgreen}3.419 & \cellcolor{bestgreen}3.710 & 3.710 & 3.677 & 3.629 \\
Claude Opus 5.5 & 3.226 & \cellcolor{bestgreen}3.710 & \cellcolor{bestgreen}3.839 & \cellcolor{bestgreen}3.774 & \cellcolor{bestgreen}3.637 \\
Gemini 4 Argon & 3.156 & 3.438 & 3.750 & 3.656 & 3.500 \\
\midrule
\rowcolor{groupblue}
\multicolumn{6}{c}{\textit{\textbf{Kling 4.0 Omni}}} \\
Direct generation & 3.394 & 3.424 & 3.758 & 3.606 & 3.545 \\
GPT-6-Astra & 3.355 & 3.742 & \cellcolor{bestgreen}3.839 & 3.613 & 3.637 \\
GPT-6.1-Sol & 3.323 & 3.742 & 3.742 & 3.613 & 3.605 \\
Claude Opus 5.5 & 3.241 & \cellcolor{bestgreen}3.793 & 3.724 & 3.621 & 3.595 \\
Gemini 4 Argon & \cellcolor{bestgreen}3.476 & 3.762 & 3.810 & \cellcolor{bestgreen}3.714 & \cellcolor{bestgreen}3.690 \\
\bottomrule
\end{tabular*}
\end{table}

\subsection{Main Results}
\label{sec:exp_main}

\textbf{\name leads on both instruction following and scene consistency.} \name achieves the highest scores across the reported VTR and R2M metrics and improves physical plausibility over direct generation. On VTR, it surpasses the strongest agentic baseline in checklist satisfaction and reduces WER from 59.21\% for direct generation to 11.08\%. These gains also extend to preserving scene structure and object states across views.

\textbf{Explicit scene modeling strengthens physical fidelity.} Each code2video method outperforms direct generation on World Models' Last Exam in Physics. \name ranks second to Claude + Blender, supporting the use of executable scenes for physical interactions. While claude also outperforms Codex with the same Blender tools, suggesting that the coding model's ability to implement physical behavior contributes to the remaining performance gap.

\textbf{Adaptive routing improves visual quality.} On T2AV-Compass, \name leads the code2video methods in realism and instruction following, with realism approaching direct generation and conventional agentic methods. This suggests that direct video generation of complex body motion reduces stiffness from proxy animation.

\begin{figure*}[t]
\centering
\includegraphics[width=\textwidth]{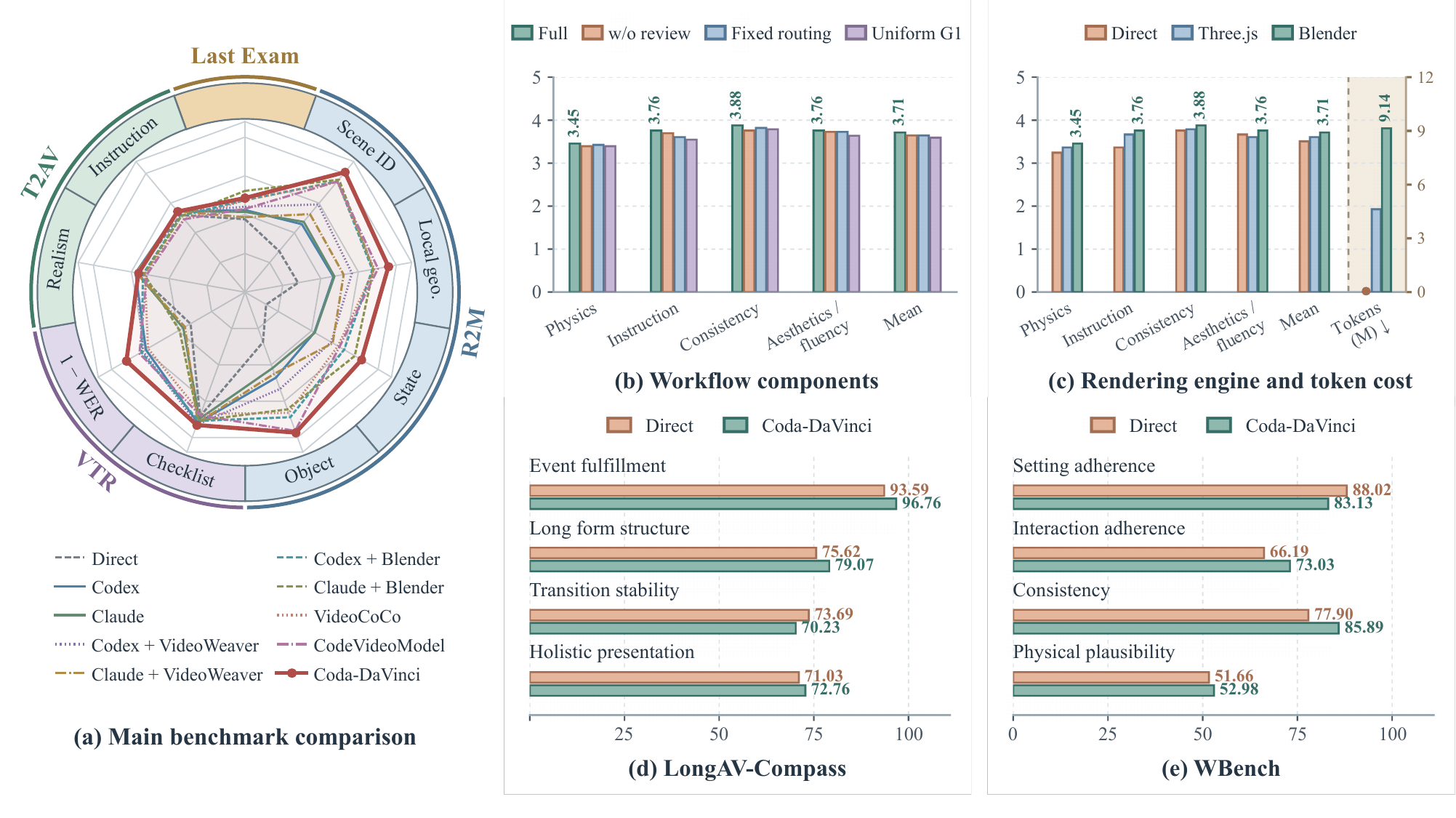}
\caption{\textbf{Performance and ablation studies.}
(a) Normalized benchmark results.
(b) Ablations of adaptive routing, independent review, and modeling granularity.
(c) Video quality and token cost across rendering engines.
(d) On LongAV-Compass, \name improves event fulfillment and long-form structure over direct generation, with lower transition stability.
(e) On WBench, it improves interaction adherence and consistency, with lower setting adherence.}
\label{fig:experiment_results}
\end{figure*}

\subsection{Ablation Experiments}
\label{sec:exp_ablation}
We conduct the ablations on a 33-task diagnostic subset. Gemini 3.8 Flash scores each video from 1 to 5 on four dimensions: physical plausibility, instruction following, visual consistency, and joint aesthetics and fluency. All configurations use the same task-specific checklists, derived from the original instructions and references. We average the four scores to obtain the overall score.

\textbf{\name is robust across coding agents.}
We evaluate \name with four coding agents and three video generators in Table~\ref{tab:agent_model_ablation}. Overall scores remain similar across coding agents, suggesting that our task-adaptive design works with different models. The video generator also affects the result: as shown in Figure~\ref{fig:proxy_adherence_cult0041}, even with correct proxy actions or camera movements, the video may deviate if the model does not follow the reference.

\textbf{Task adaptation and independent review improve video quality.}
We ablate three components in \name with GPT-6-Astra and Seedance 2.5: independent acceptance, adaptive routing, and entity-wise modeling granularity. Each ablation lowers scores on all four dimensions as shown in Figure~\ref{fig:experiment_results}(b). Independent acceptance is particularly useful for consistency, which drops most when this review is removed.
Fixing routing  to 3D modeling in all tasks or applying \texttt{G1} to all subjects has its largest effect on instruction following and aesthetics. These results suggest that fulfilling the task depends on both when to use explicit modeling and how much detail to represent.

\textbf{Blender improves video quality at a higher token cost.}
We replace Blender with Three.js while keeping the coding agent and other settings fixed. As shown in Figure~\ref{fig:experiment_results}(c), both modeling engines improve overall quality over direct generation. Blender scores higher on all four dimensions, suggesting that it provides a better basis for representing these tasks. It also consumes roughly twice as many agent tokens per case. Three.js thus offers a less costly option, with a modest loss in video quality on this subset.

\subsection{Long-Video Consistency Study}
\label{sec:exp_continuation}

A reusable 3D scene preserves objects and spatial relationships across clips, providing persistent memory for long-video generation.

We compare \name with direct Seedance 2.5 on 29 LongAV-Compass and 50 WBench cases. LongAV-Compass sequences span 25--90 seconds and are generated event by event, using the preceding tail frame (T2AV/I2AV) or clip (V2AV) as reference. WBench evaluates successive interactions within a scene, with each turn continuing from the previous tail frame.

As shown in Figure~\ref{fig:experiment_results}(d,e), \name improves long-form structure on LongAV-Compass, with its largest WBench gains in interaction adherence and consistency. These findings support scene reuse for maintaining state across successive instructions. However, direct generation performs better in transition stability on LongAV-Compass and setting adherence on WBench, revealing remaining limitations in cross-clip continuity and scene fidelity.

\subsection{Qualitative Analysis}
\label{sec:exp_qualitative}

\begin{figure}[!htbp]
\centering
\includegraphics[width=\linewidth]{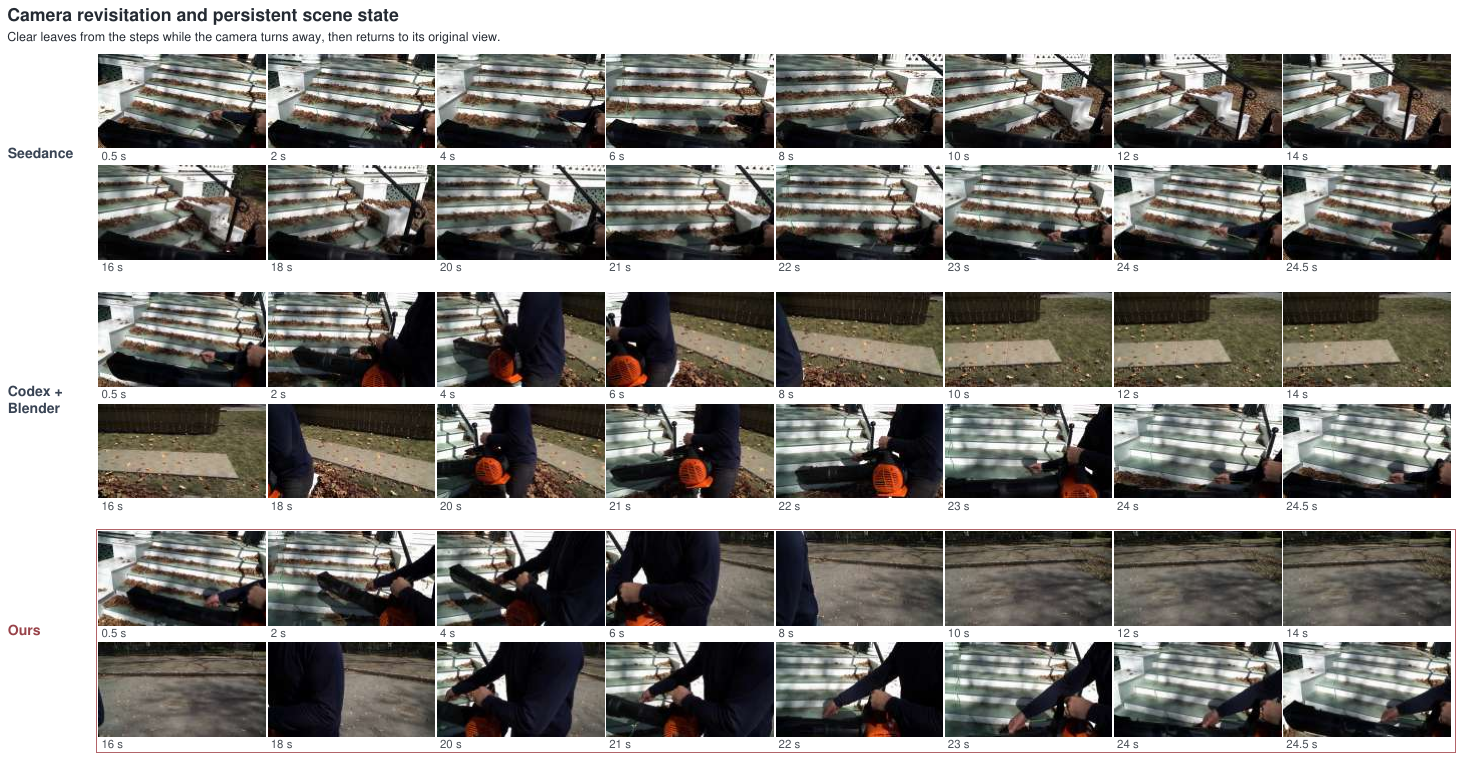}
\caption{\textbf{Camera instruction following and state persistence.} The camera turns away while leaves are being cleared, then returns to the steps. Each method occupies two rows, read consecutively from left to right. Frames are sampled at matching timestamps, with denser sampling during the return. Both code2video methods show cleared steps on revisitation, while Seedance retains substantial leaf coverage.}
\label{fig:qualitative_revisit}
\end{figure}

\textbf{Scene modeling supports instruction following and state persistence.}
In Figure~\ref{fig:qualitative_revisit}, the camera is instructed to turn away from a person clearing leaves and then return to the steps. Seedance keeps the steps largely in view, with many leaves still present near the end. Both Codex + Blender and \name complete the camera departure and return to cleared steps. The ongoing action therefore advances across the unseen interval and leaves a lasting change in the scene. This example illustrates how explicit modeling helps coordinate camera instructions with event progression while preserving the resulting state across a revisit.

\begin{figure}[!htbp]
\centering
\includegraphics[width=\linewidth]{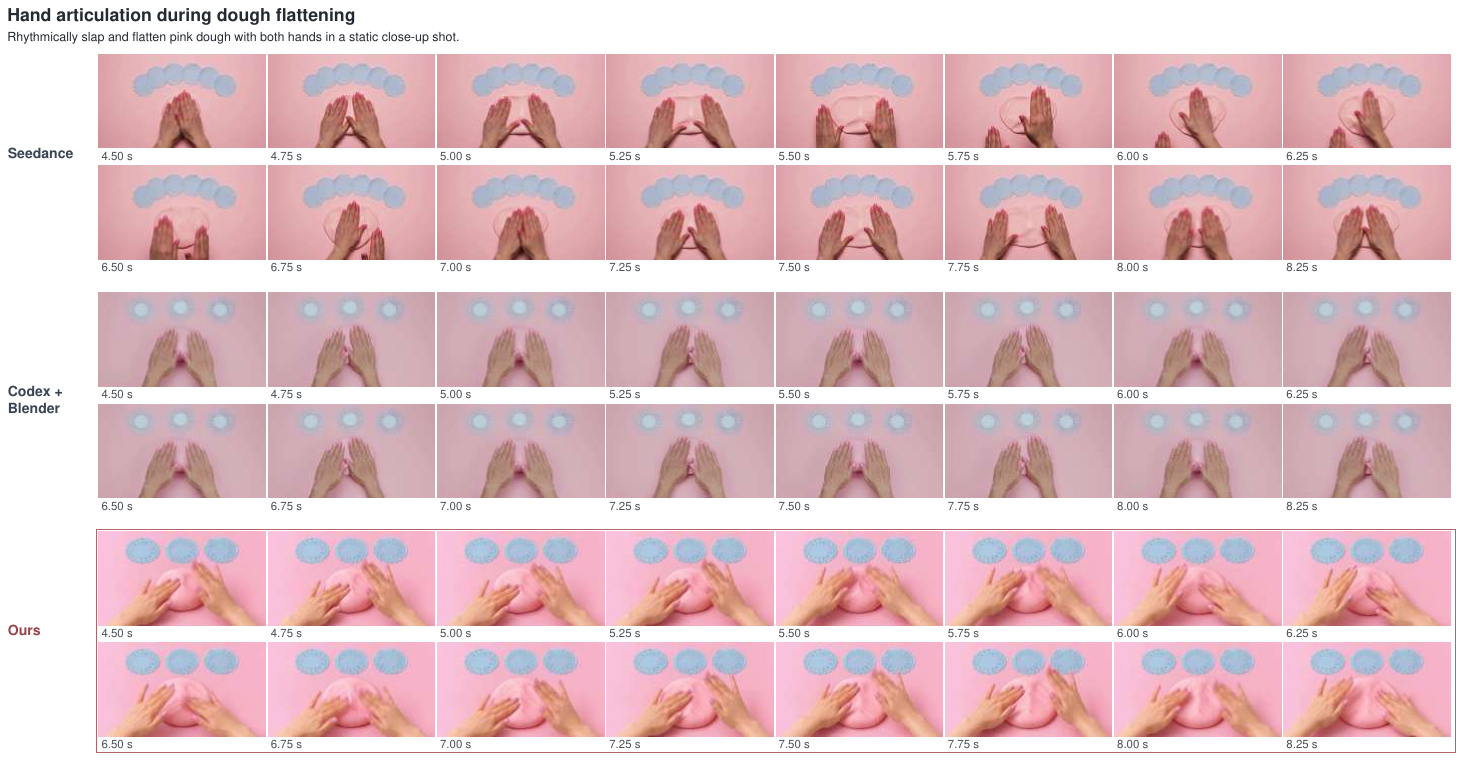}
\caption{\textbf{Hand articulation and visual fluency.} Consecutive frames from a dough-flattening task, sampled at 4 fps over 4.50--8.25 seconds. Each method occupies two rows, with the second continuing the first. Compared with the nearly fixed pose of Codex + Blender, \name shows changing wrist and finger poses through repeated contact with the dough. Seedance provides the direct-generation comparison.}
\label{fig:qualitative_motion}
\end{figure}

\textbf{Selective modeling preserves natural hand motion.}
Figure~\ref{fig:qualitative_motion} compares a close-up of hands flattening dough. Codex + Blender keeps the hands in nearly the same pose throughout the sequence, giving the interaction a static appearance. In \name, the wrists turn and the fingers adjust through successive presses as the dough changes shape. These coordinated changes make the action more fluid and the contact more convincing, retaining the lively motion visible in native Seedance generation. Here, \name uses the Blender route: the result illustrates the value of leaving fine articulation to the video model while using the proxy to organize the interaction.

\section{Conclusion}
\label{sec:conclusion}

We investigated how coding agents and video-generation models can be coordinated within the Code2Video paradigm. Our analysis shows that effective proxy guidance requires matching modeling granularity, physics simulation, and prompting to task requirements. These findings inform \name{}, which assigns essential structure and dynamics to coding agents while leaving appearance and fine motion to video models, supported by persistent scene assets and visual feedback. Experiments demonstrate leading performance on VTR and R2M, improved physical plausibility over direct generation, and further gains in long-form structure and consistency across successive interactions. Remaining gaps in motion naturalness, transition stability, and adherence to scene settings highlight the need to better preserve explicit control during synthesis as video generation extends to longer and more complex scenarios.
\begingroup
\setlength{\bibsep}{3pt plus 1pt minus 1pt}
\bibliographystyle{plainnat}
\bibliography{main}
\endgroup

\clearpage
\appendix
\section{Case Study}
\label{app:case_study}

We compare the generated videos with the proxies supplied to the video model. Each example shows the same task across the evaluated configurations, using shared timestamps to follow the scene through the sequence.

\subsection{Scene and State Consistency}

\begin{figure}[!htbp]
\centering
\includegraphics[page=1,width=\linewidth]{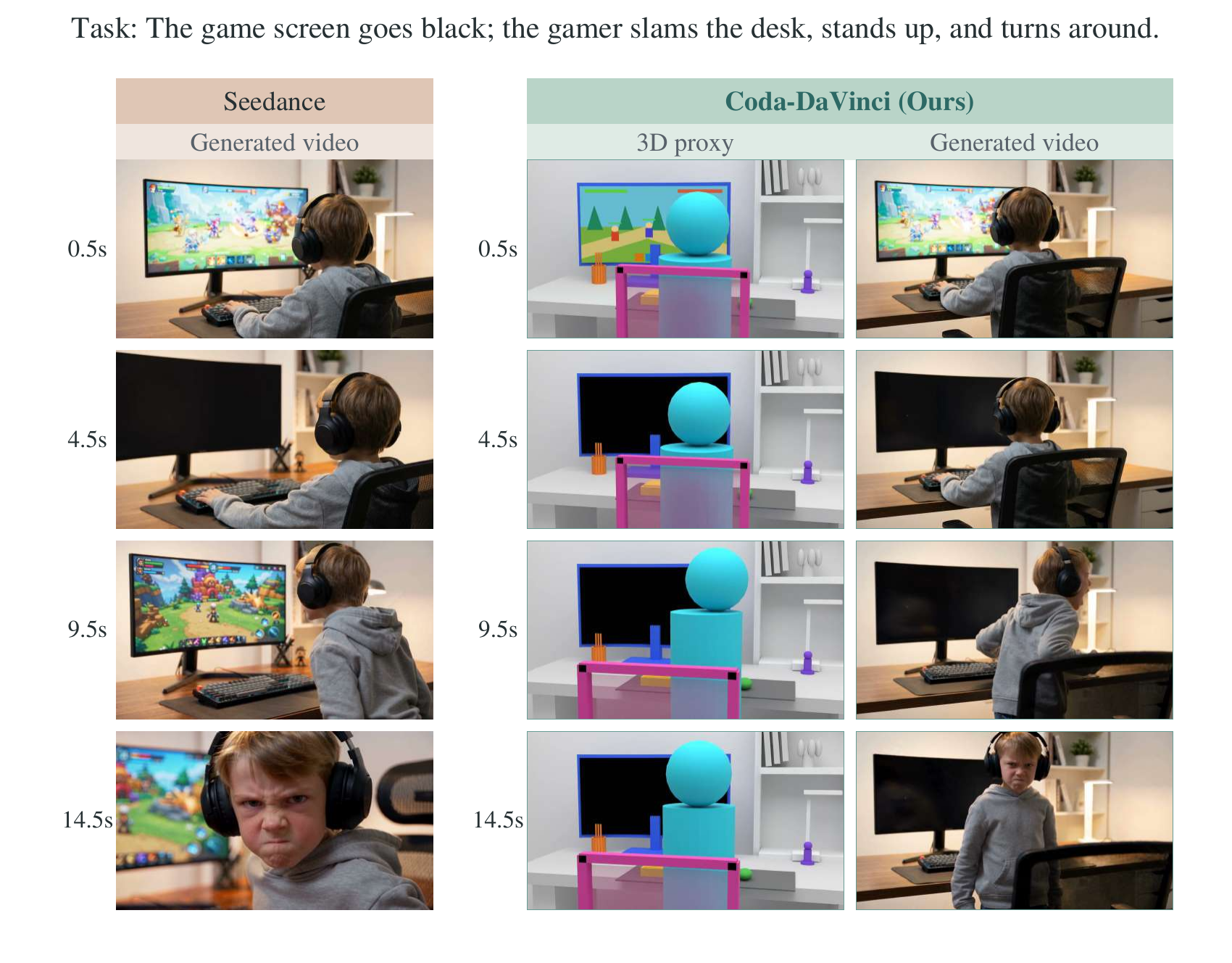}
\caption{\textbf{Preserving an edited scene state.} WBench case 248. The monitor goes black before the player stands up and turns around. Direct generation brings back the game display in later turns, while \name preserves the black-screen state in both the proxy and the generated video.}
\label{fig:case_black_screen}
\end{figure}

\begin{figure}[!htbp]
\centering
\includegraphics[page=1,width=\linewidth]{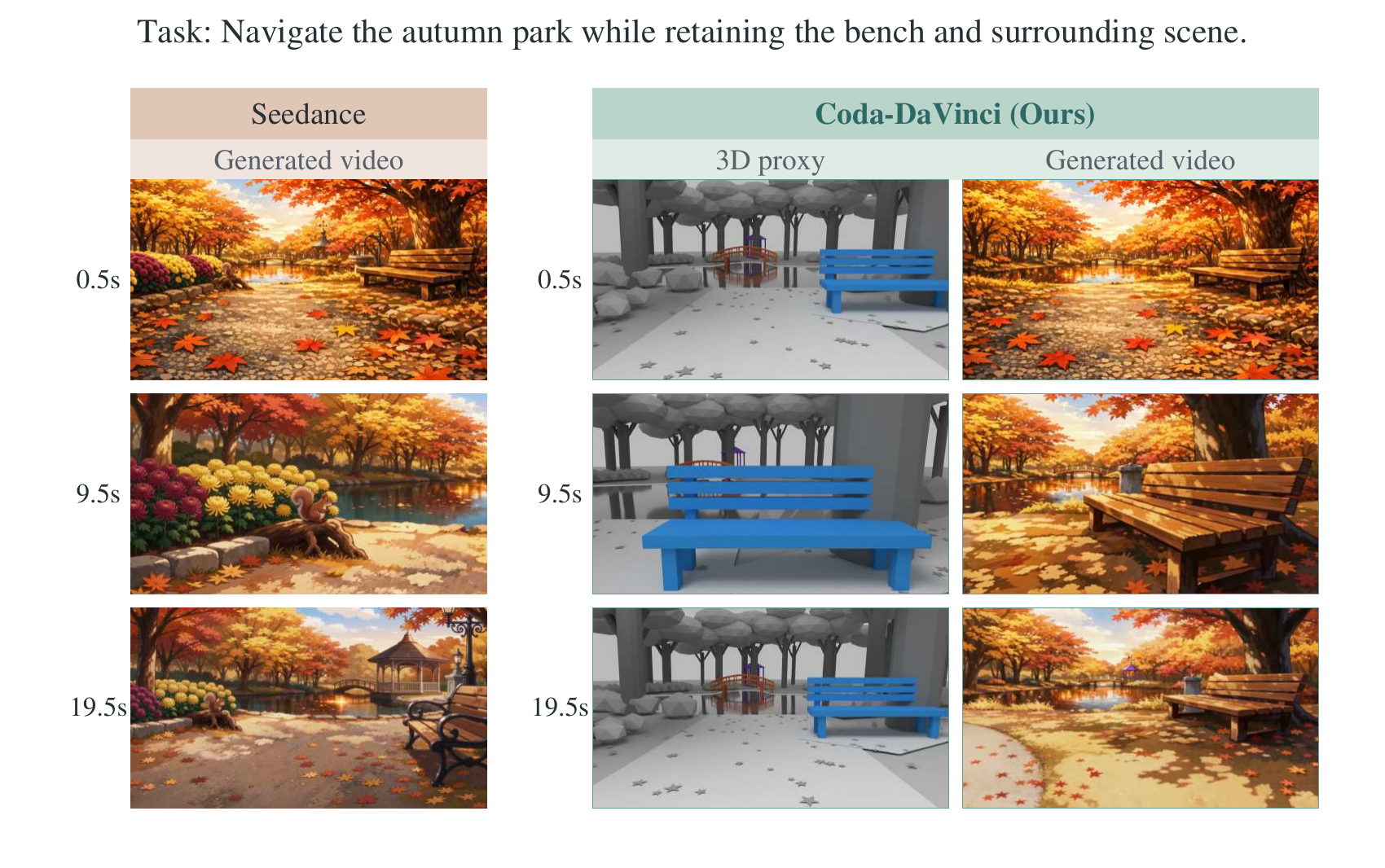}
\caption{\textbf{Scene consistency across successive views.} WBench case 92. The camera moves through an autumn park. \name retains the wooden bench across views; the direct-generation sequence introduces a bench with a different armrest structure. Corresponding proxy frames show the shared scene used for continuation.}
\label{fig:case_park_bench}
\end{figure}

The transit-plaza example in Figure~\ref{fig:case_r2m_plaza} shows how the circular portal and overhead structure persist through camera motion. State changes also need to survive later instructions: in Figure~\ref{fig:case_black_screen}, \name keeps the monitor black as the player reacts, whereas direct generation restores the game image. Figure~\ref{fig:case_park_bench} illustrates the same issue across views, with the modeled bench retaining its structure as the camera moves through the park.

\begin{figure}[!htbp]
\centering
\includegraphics[page=1,width=\linewidth]{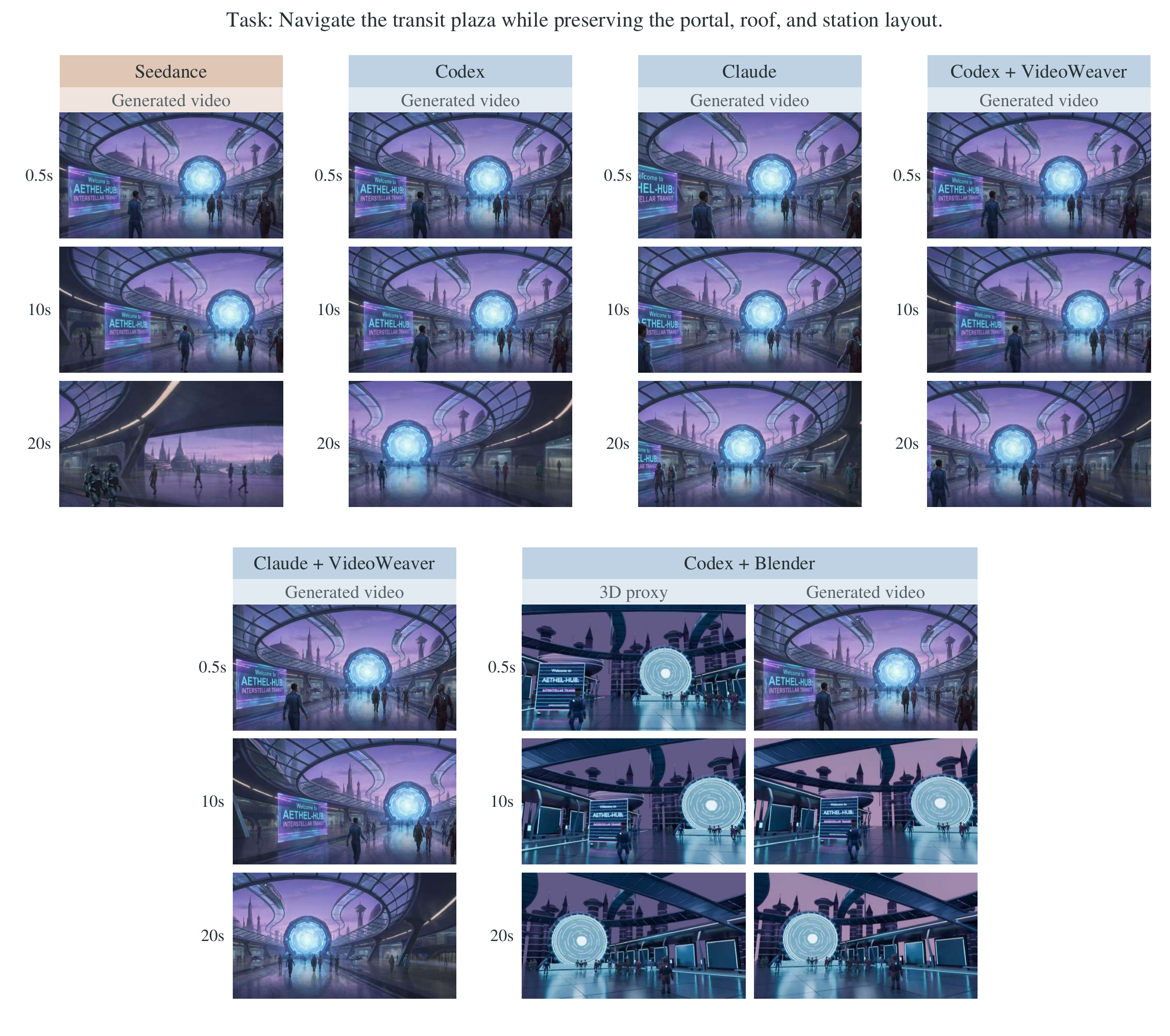}
\caption{\textbf{Scene geometry during navigation.} R2M-Bench case \texttt{WSLRRLL\_0046}. The portal, curved roof, and surrounding station provide landmarks for comparing scene structure through the camera movement. Proxy renders and generated frames reveal how each modeled scene carries through synthesis.}
\label{fig:case_r2m_plaza}
\end{figure}
\clearpage

\begin{figure}[!htbp]
\ContinuedFloat
\centering
\includegraphics[page=2,width=\linewidth]{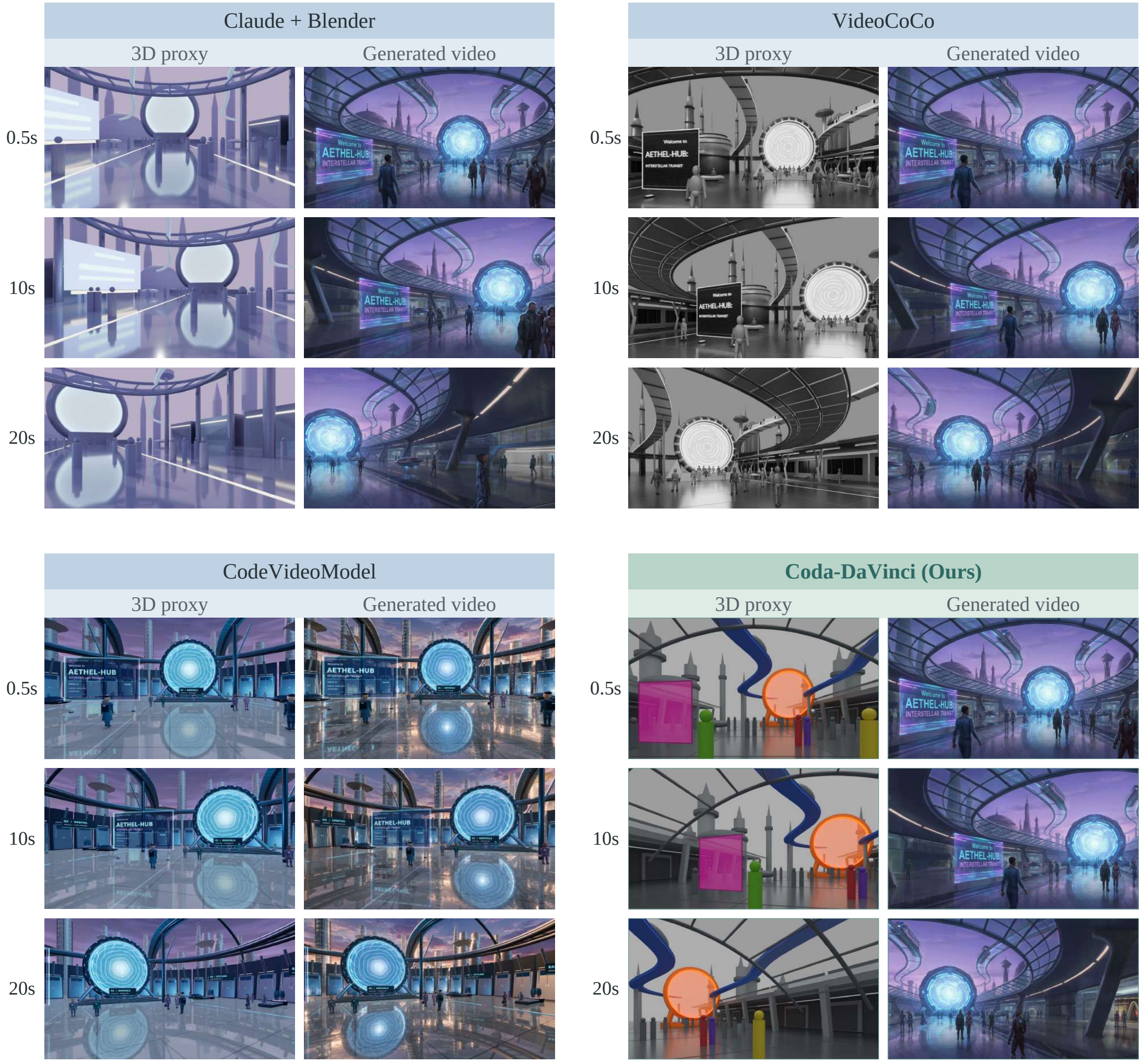}
\caption{\textbf{Scene geometry during navigation (continued).} The remaining configurations are shown at the same timestamps, with their corresponding proxy renders.}
\end{figure}
\clearpage

\clearpage

\clearpage

\subsection{Instruction Following}

We next examine tasks that combine several objects and actions. The cassette and keyboard examples in Figures~\ref{fig:case_cassette} and~\ref{fig:case_keyboard} show how explicit object layout supports the requested interactions while leaving room for natural human appearance. The pet-care example in Figure~\ref{fig:case_pet_care} adds a camera constraint: \name presents the handover and subsequent actions within a continuous wide view.

\begin{figure}[!htbp]
\centering
\includegraphics[page=1,width=\linewidth]{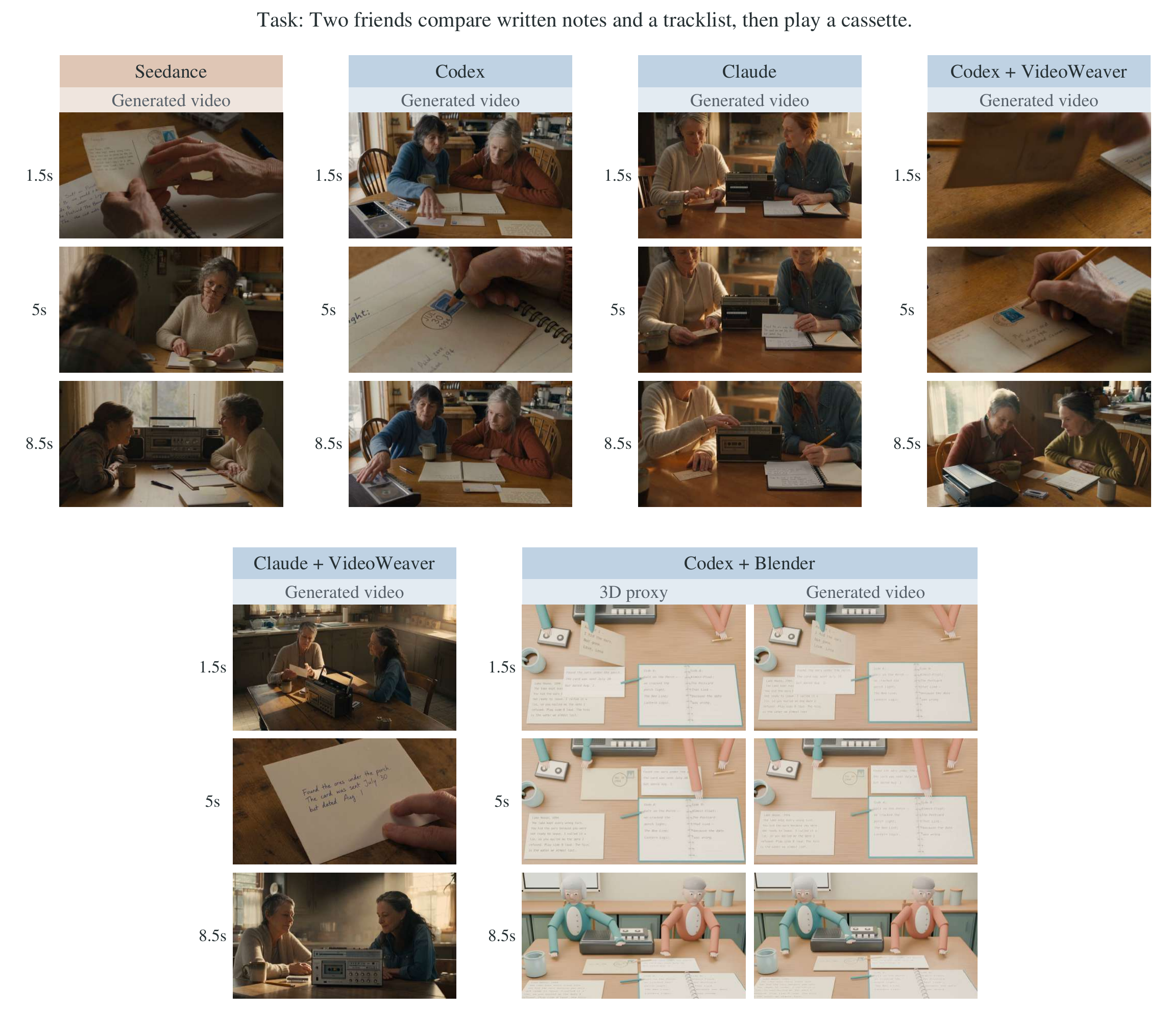}
\caption{\textbf{Coordinating objects and interactions.} VTR-Bench case \texttt{CULT-0034}. Two friends compare a set of written notes and a tracklist before playing a cassette. \name represents the tabletop objects and subject identities in the proxy, while the final video supplies natural hand movements and human appearance.}
\label{fig:case_cassette}
\end{figure}
\clearpage

\begin{figure}[!htbp]
\ContinuedFloat
\centering
\includegraphics[page=2,width=\linewidth]{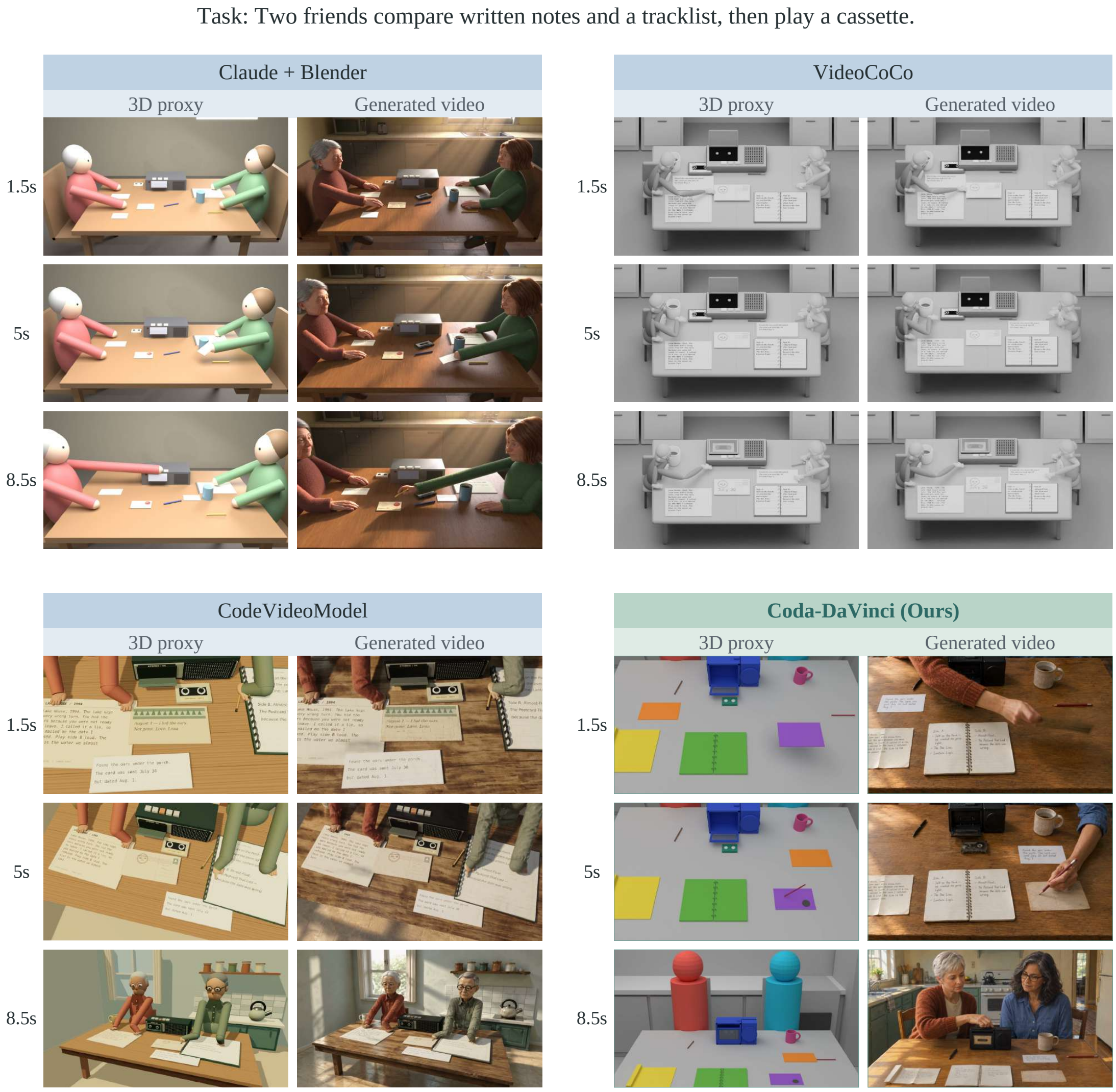}
\caption{\textbf{Coordinating objects and interactions (continued).} The remaining configurations are shown at the same timestamps, with their corresponding proxy renders.}
\end{figure}
\clearpage

\begin{figure}[!htbp]
\centering
\includegraphics[page=1,width=\linewidth]{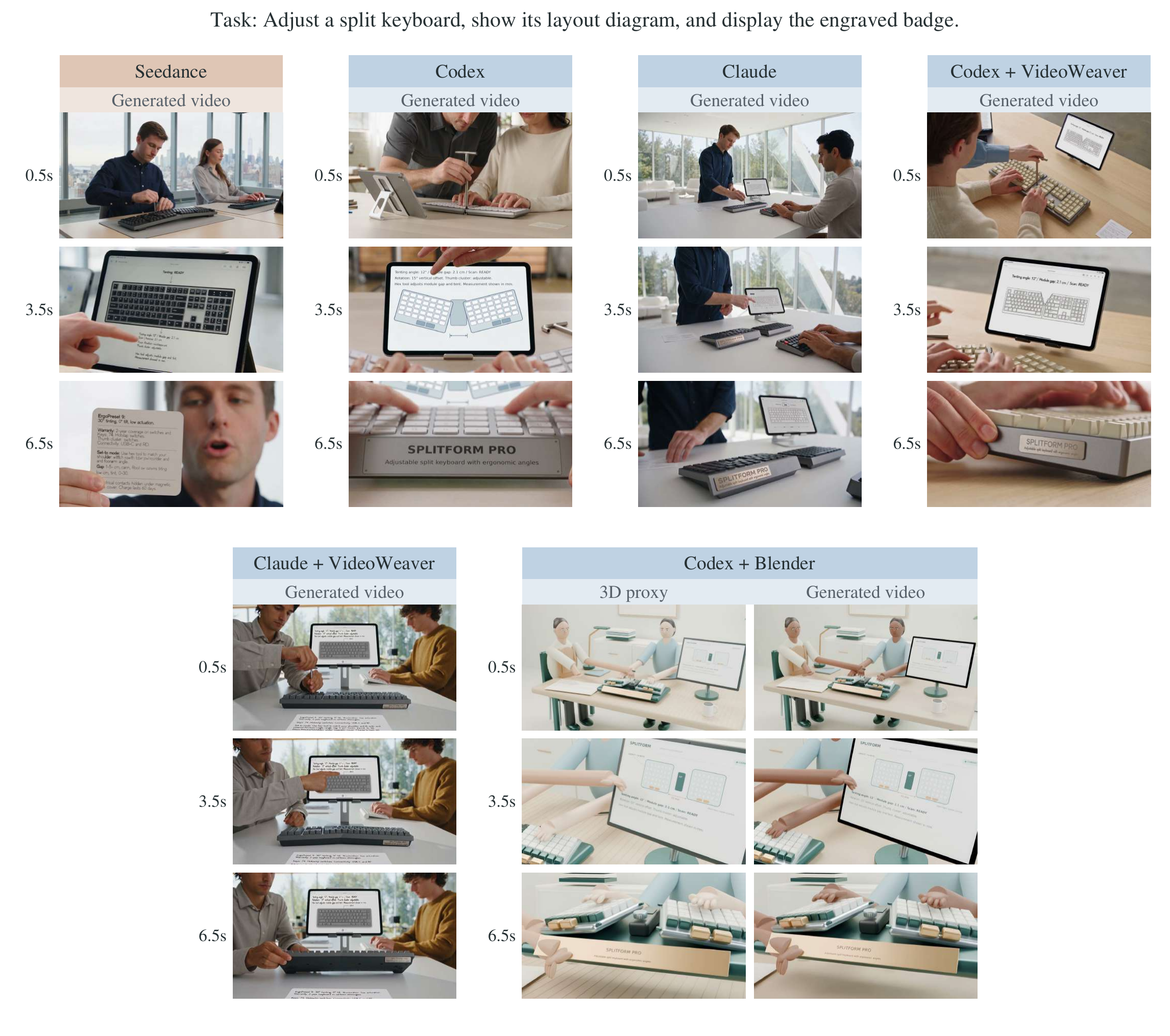}
\caption{\textbf{Preserving task-critical object structure.} VTR-Bench case \texttt{AD-0059}. The task involves adjusting a split keyboard, inspecting its layout diagram, and showing its engraved badge. The proxy specifies the keyboard components and camera views, which guide the corresponding actions in the generated video.}
\label{fig:case_keyboard}
\end{figure}
\clearpage

\begin{figure}[!htbp]
\ContinuedFloat
\centering
\includegraphics[page=2,width=\linewidth]{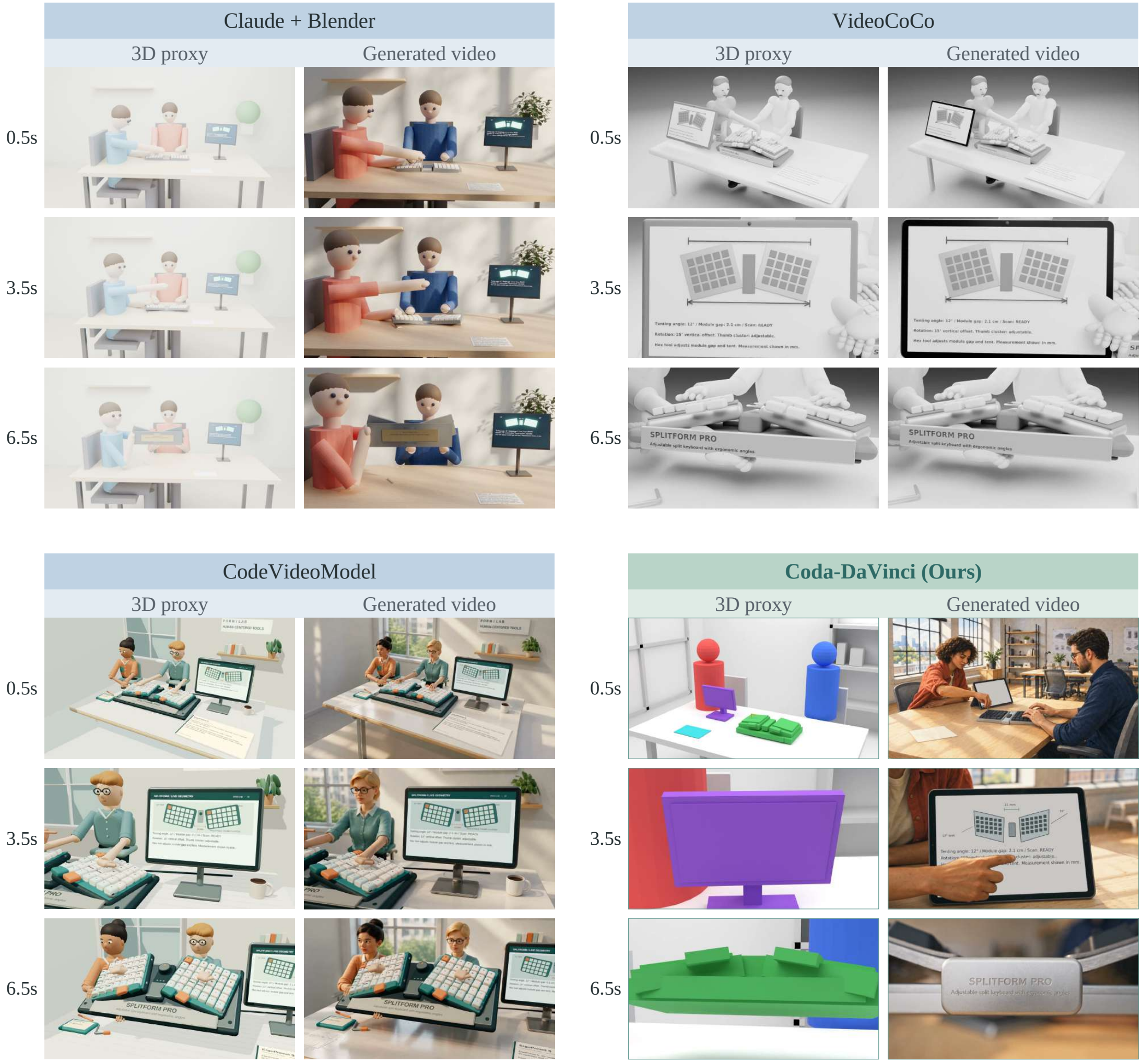}
\caption{\textbf{Preserving task-critical object structure (continued).} The remaining configurations are shown at the same timestamps, with their corresponding proxy renders.}
\end{figure}
\clearpage

\begin{figure}[!htbp]
\centering
\includegraphics[page=1,width=\linewidth]{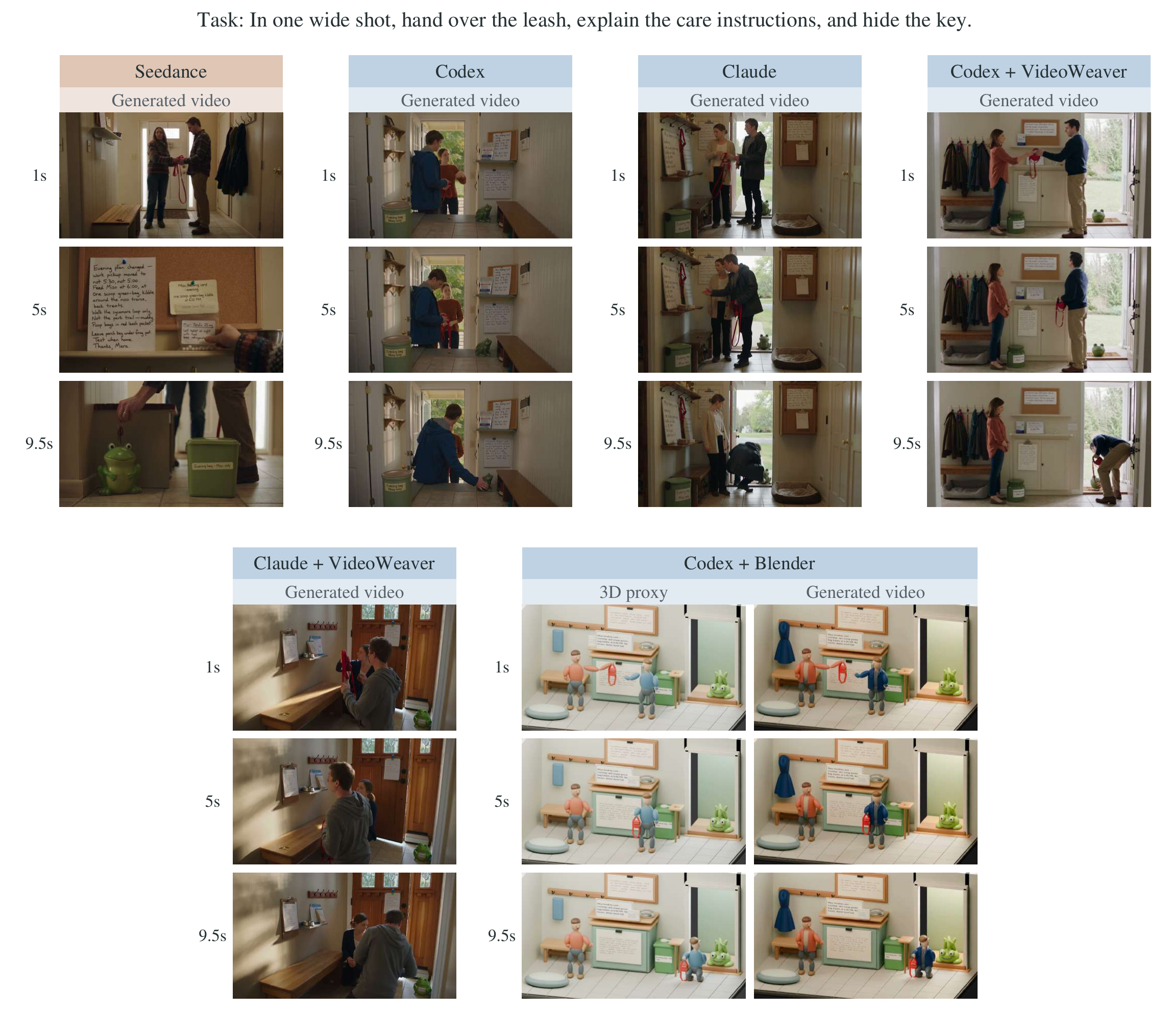}
\caption{\textbf{Following actions within a continuous shot.} VTR-Bench case \texttt{LIFE-0008}. The owner hands over a leash and explains the care instructions before the neighbor hides a key under the frog-shaped planter. \name keeps these actions within the requested wide view, preserving the room layout throughout.}
\label{fig:case_pet_care}
\end{figure}
\clearpage

\begin{figure}[!htbp]
\ContinuedFloat
\centering
\includegraphics[page=2,width=\linewidth]{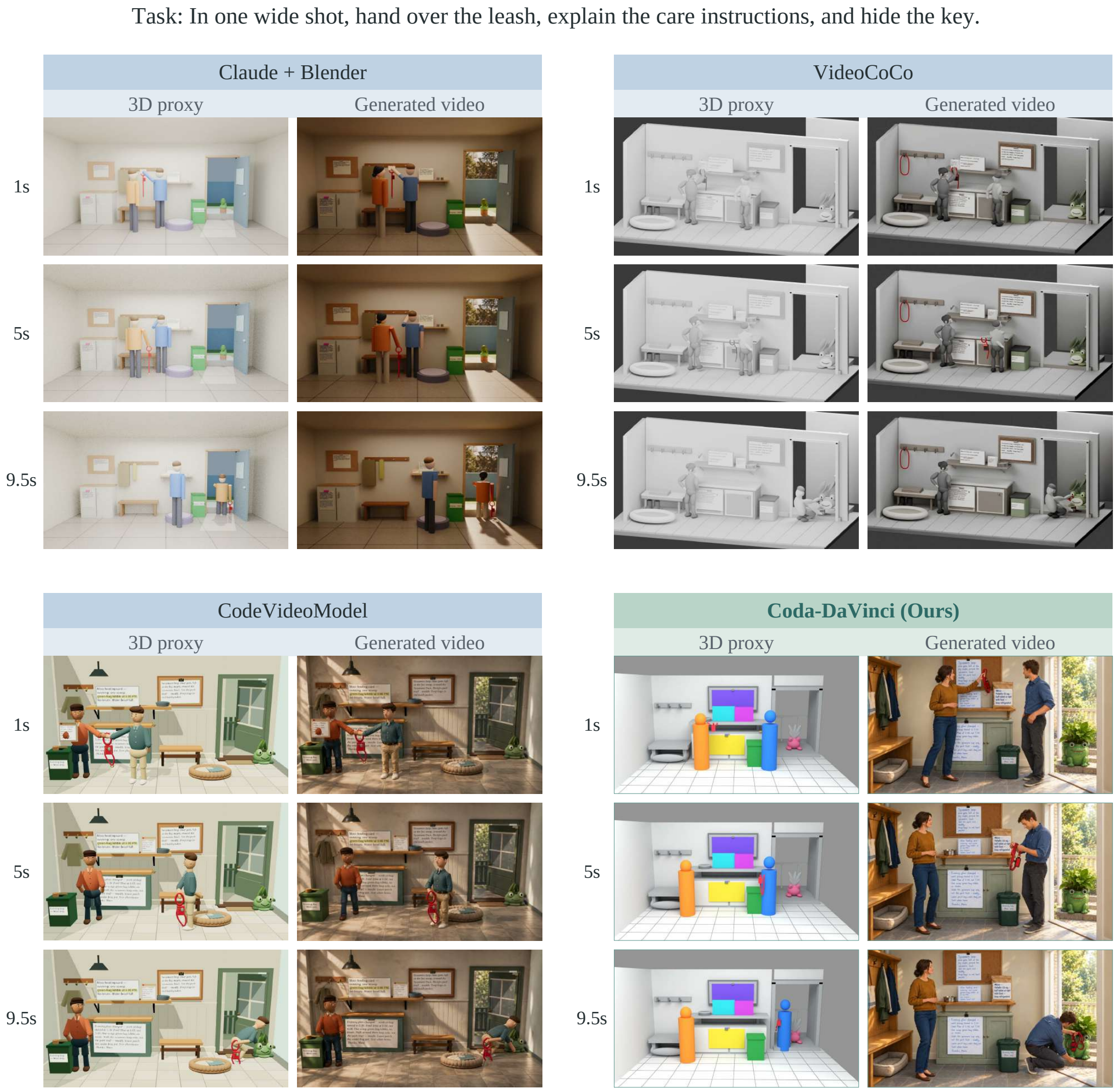}
\caption{\textbf{Following actions within a continuous shot (continued).} The remaining configurations are shown at the same timestamps, with their corresponding proxy renders.}
\end{figure}
\clearpage

\subsection{Selective Modeling and Visual Quality}

The following examples show how simplified proxies can guide the scene while allowing the video model to supply finer appearance and motion. In Figure~\ref{fig:case_dough}, the proxy describes the flattening dough and surrounding objects, and the final video adds the hands performing the action. The wheelchair interaction in Figure~\ref{fig:case_wheelchair} and the alternating cockpit views in Figure~\ref{fig:case_cockpit} extend this division of work to human movement and expression. Figure~\ref{fig:case_dumplings} provides an object-focused example, tracing the transition from modeled food geometry to the final surface detail.

\begin{figure}[!htbp]
\centering
\includegraphics[page=1,width=\linewidth]{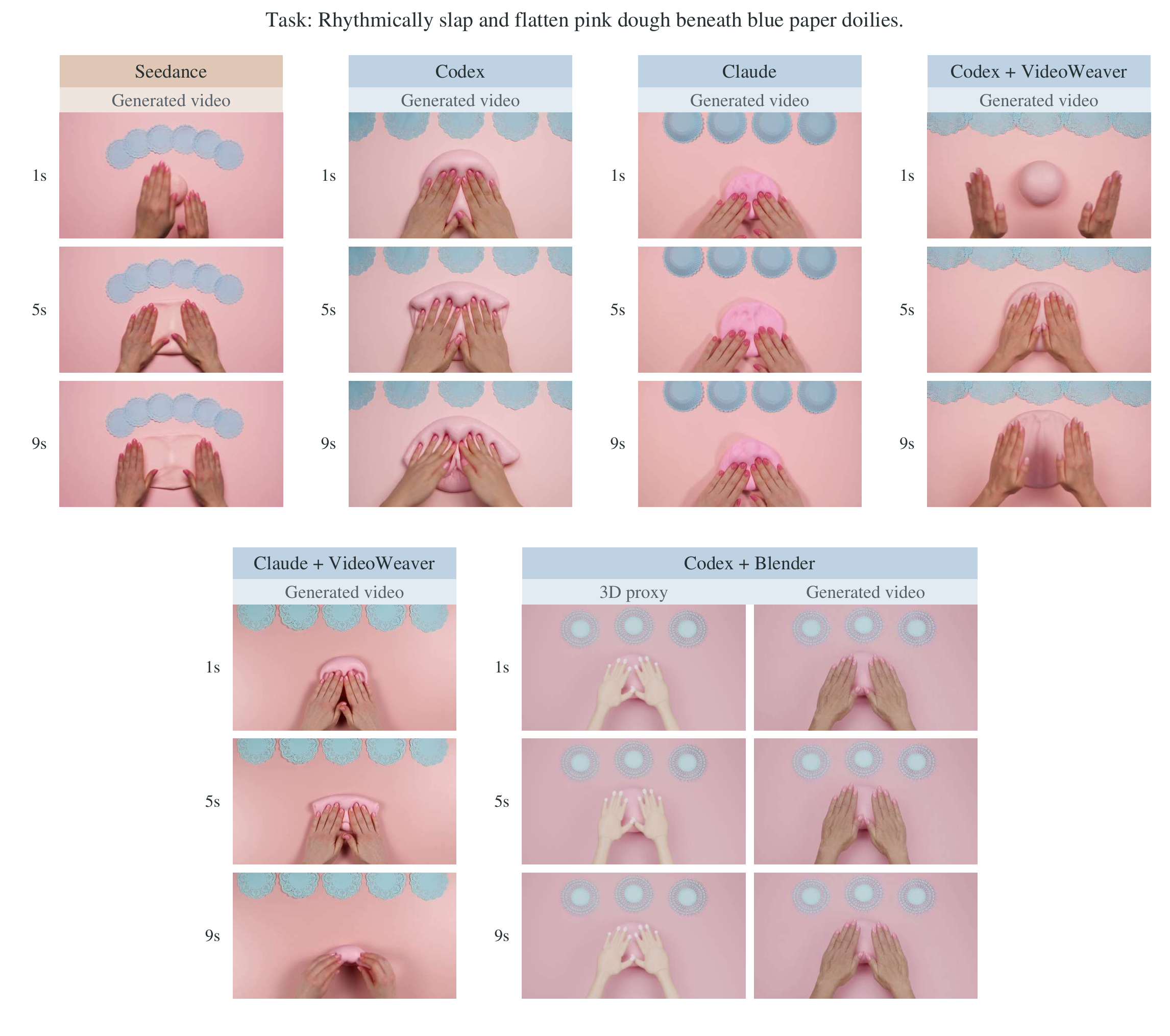}
\caption{\textbf{Deformation with natural hand motion.} T2AV-Compass case \texttt{0348}. The proxy controls the dough flattening beneath the paper doilies. \name leaves the hands to the video model, producing the requested manipulation around the changing dough shape.}
\label{fig:case_dough}
\end{figure}
\clearpage

\begin{figure}[!htbp]
\ContinuedFloat
\centering
\includegraphics[page=2,width=\linewidth]{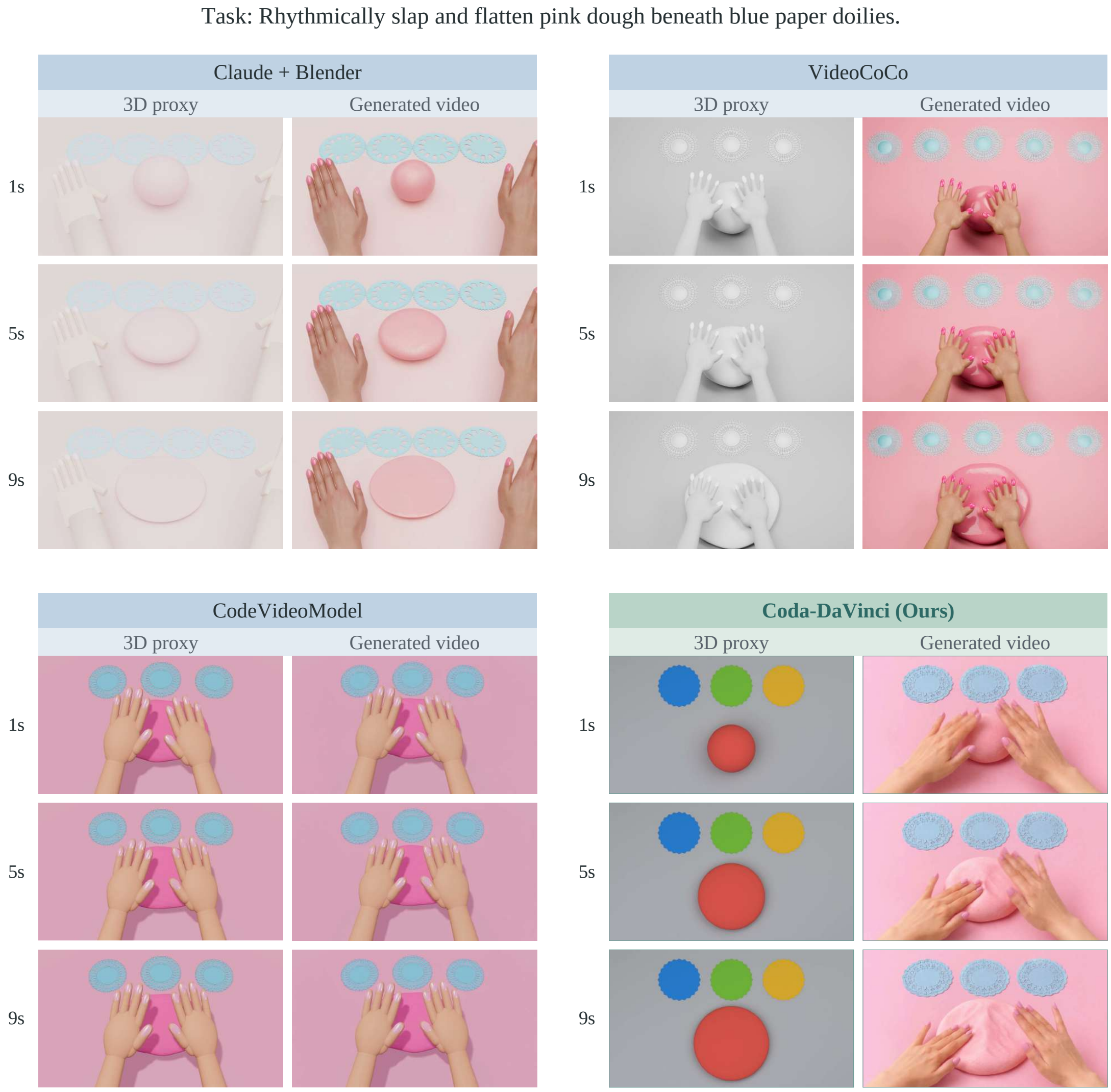}
\caption{\textbf{Deformation with natural hand motion (continued).} The remaining configurations are shown at the same timestamps, with their corresponding proxy renders.}
\end{figure}
\clearpage

\begin{figure}[!htbp]
\centering
\includegraphics[page=1,width=\linewidth]{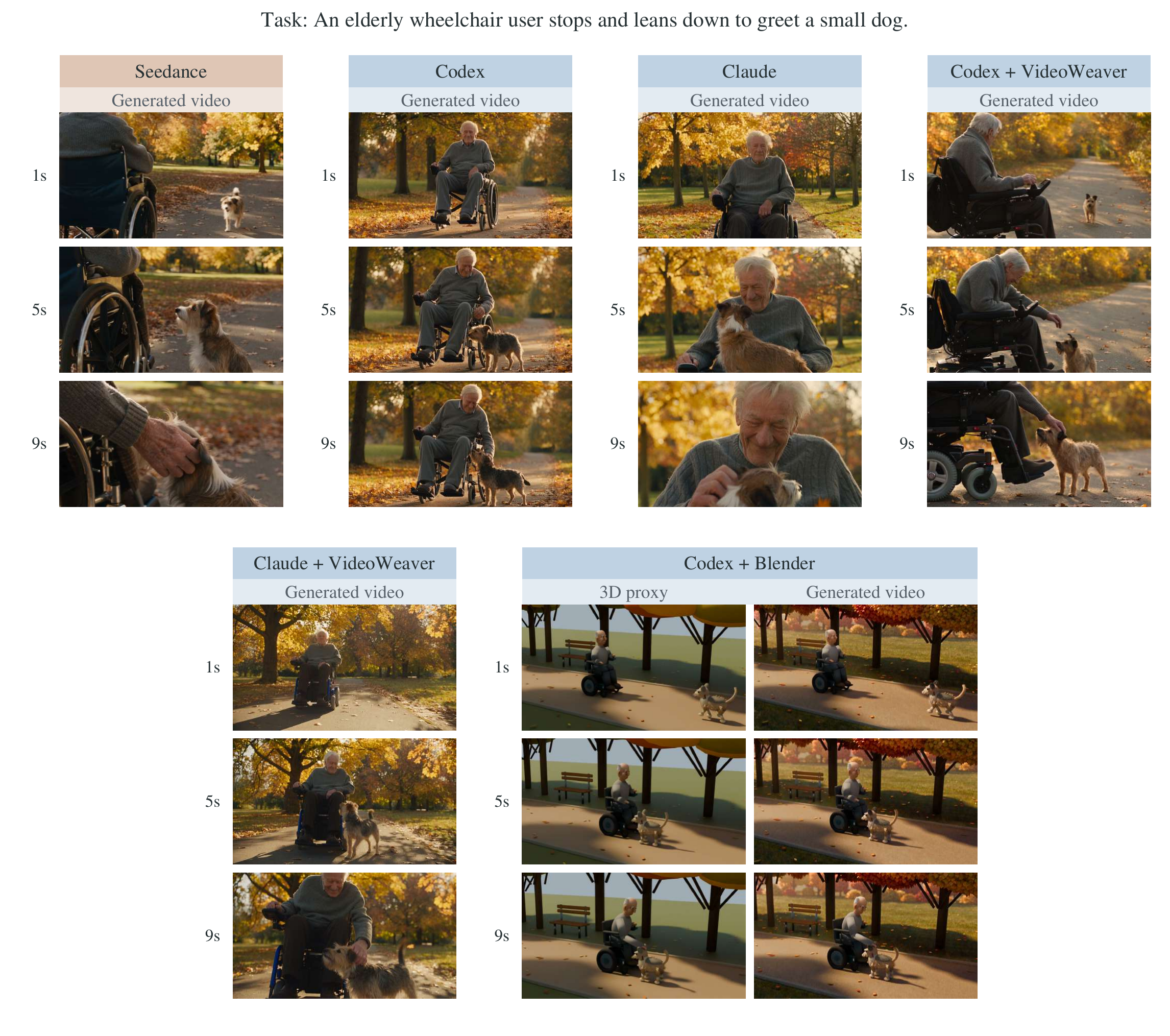}
\caption{\textbf{Natural interaction from a simplified scene.} T2AV-Compass case \texttt{0333}. An elderly wheelchair user stops to greet a small dog. The proxy establishes their relative positions and motion, while the generated video supplies the finer posture and appearance of the interaction.}
\label{fig:case_wheelchair}
\end{figure}
\clearpage

\begin{figure}[!htbp]
\ContinuedFloat
\centering
\includegraphics[page=2,width=\linewidth]{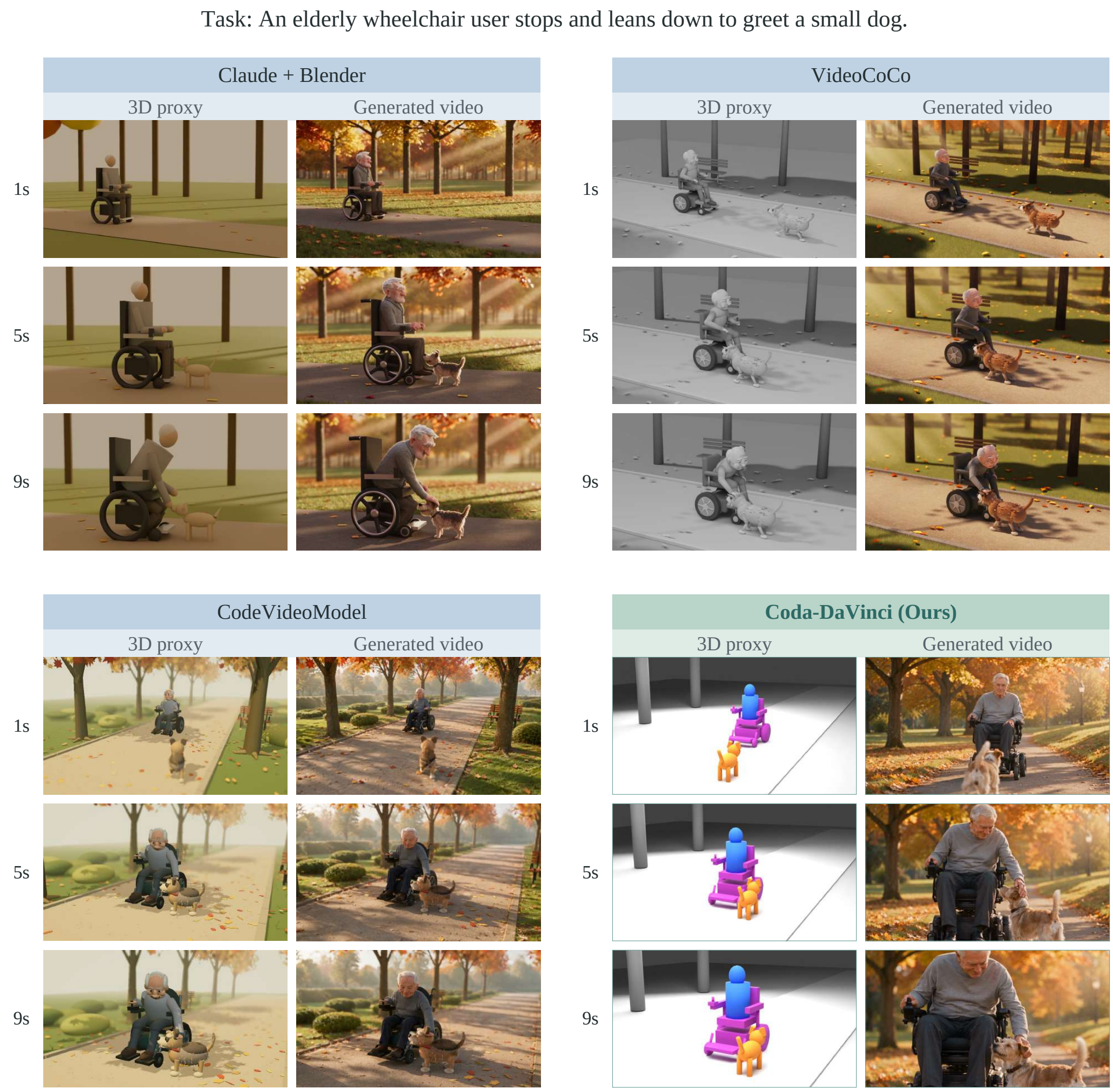}
\caption{\textbf{Natural interaction from a simplified scene (continued).} The remaining configurations are shown at the same timestamps, with their corresponding proxy renders.}
\end{figure}
\clearpage

\begin{figure}[!htbp]
\centering
\includegraphics[page=1,width=\linewidth]{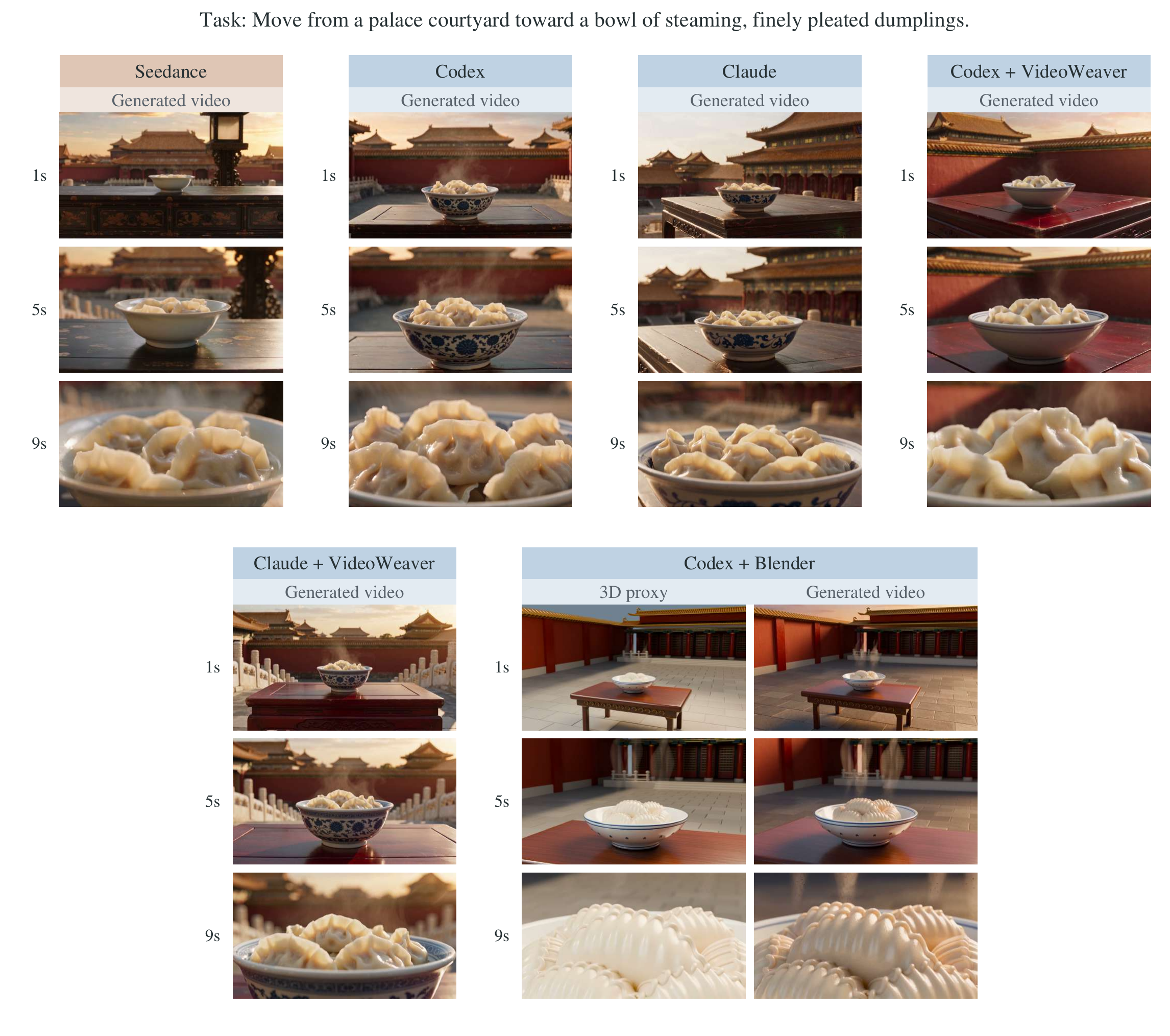}
\caption{\textbf{From modeled objects to fine visual detail.} T2AV-Compass case \texttt{0064}. The camera approaches a bowl of steaming dumplings in a palace courtyard. Paired proxy and generated frames show how the modeled arrangement is rendered into food texture, pleats, and steam.}
\label{fig:case_dumplings}
\end{figure}
\clearpage

\begin{figure}[!htbp]
\ContinuedFloat
\centering
\includegraphics[page=2,width=\linewidth]{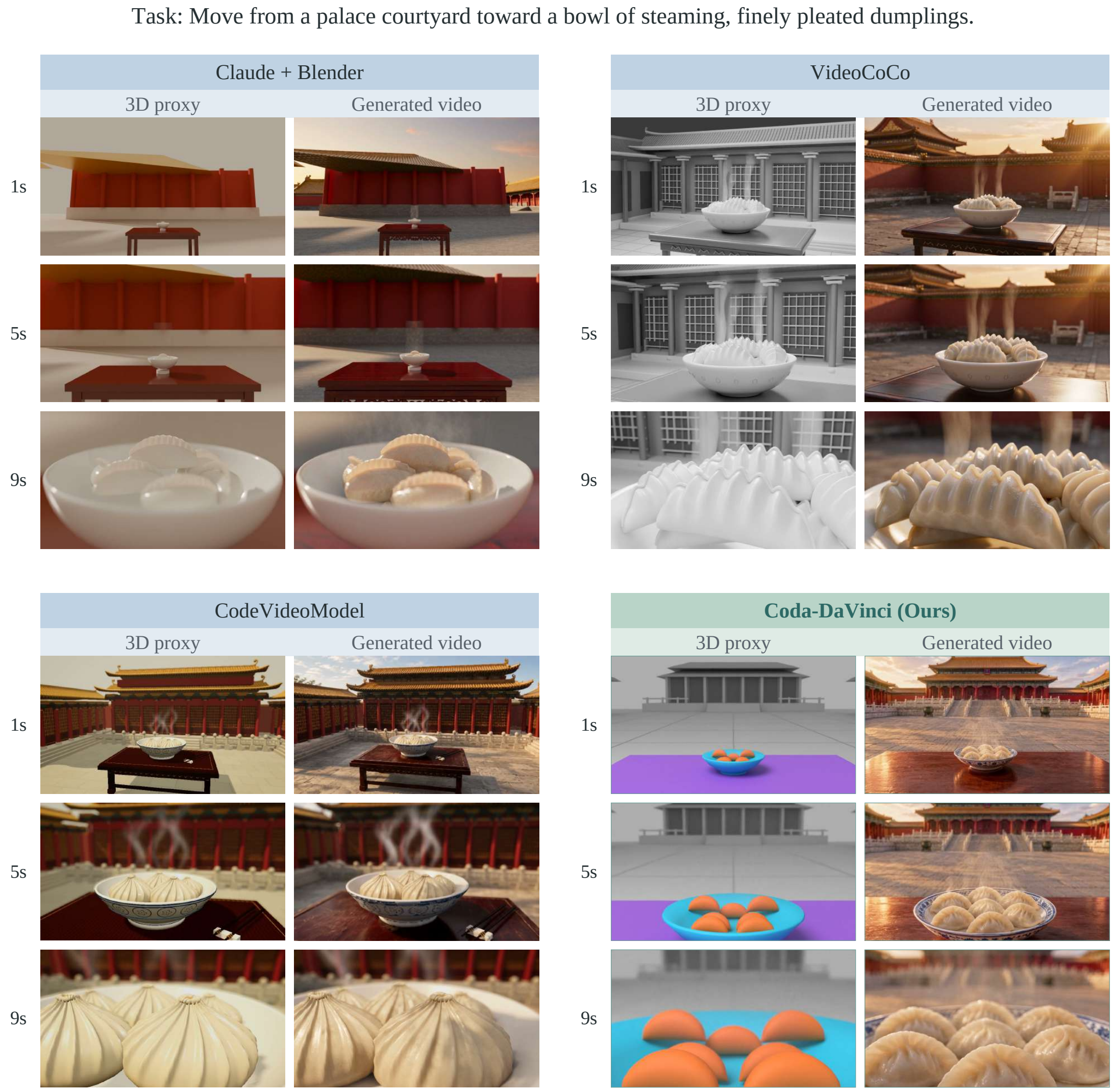}
\caption{\textbf{From modeled objects to fine visual detail (continued).} The remaining configurations are shown at the same timestamps, with their corresponding proxy renders.}
\end{figure}
\clearpage

\begin{figure}[!htbp]
\centering
\includegraphics[page=1,width=\linewidth]{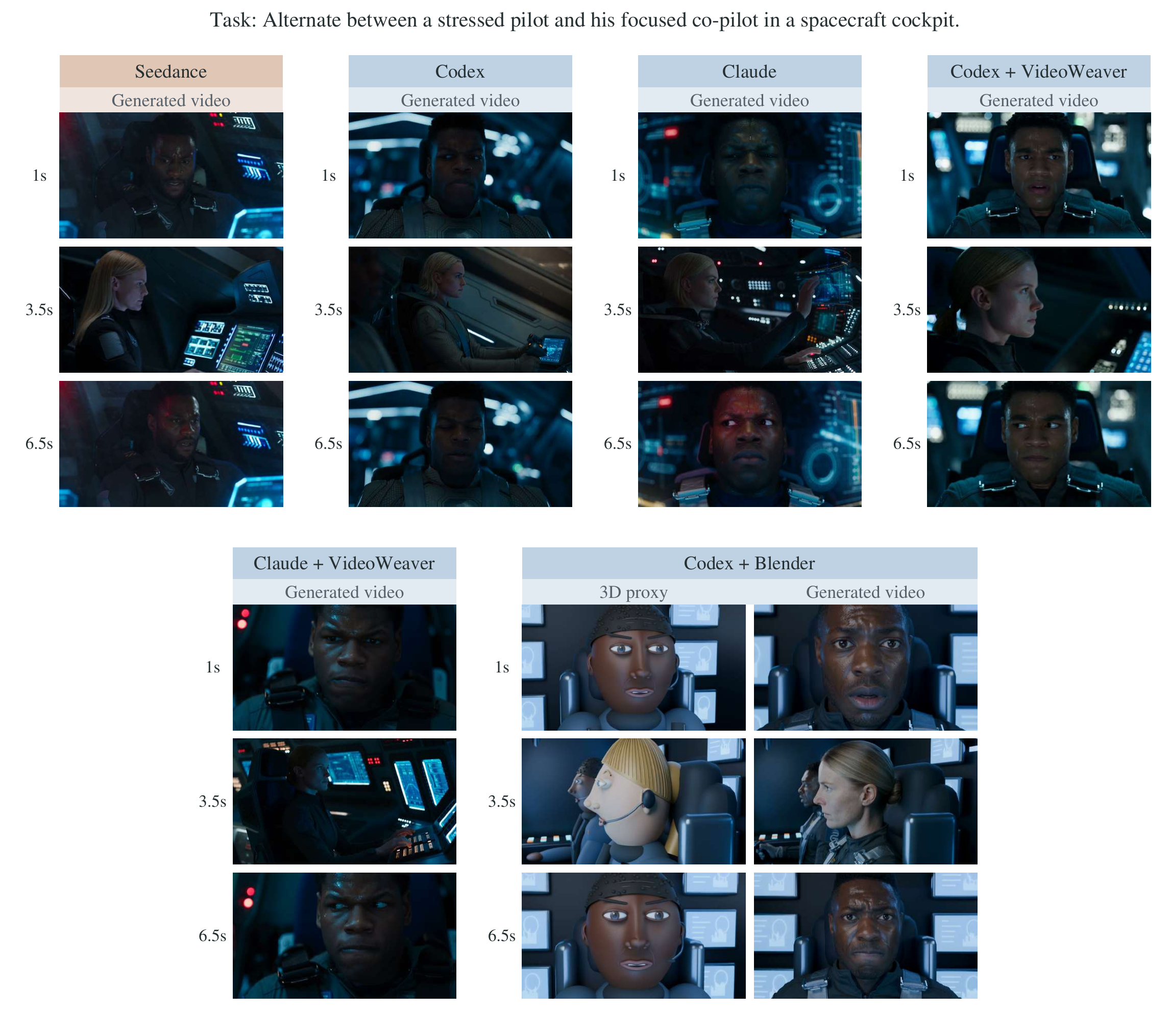}
\caption{\textbf{Subject identity across alternating close-ups.} T2AV-Compass case \texttt{0424}. The sequence alternates between a stressed pilot and his focused co-pilot. \name uses simple subject proxies to establish identity and framing, allowing the video model to generate facial detail and expression.}
\label{fig:case_cockpit}
\end{figure}
\clearpage

\begin{figure}[!htbp]
\ContinuedFloat
\centering
\includegraphics[page=2,width=\linewidth]{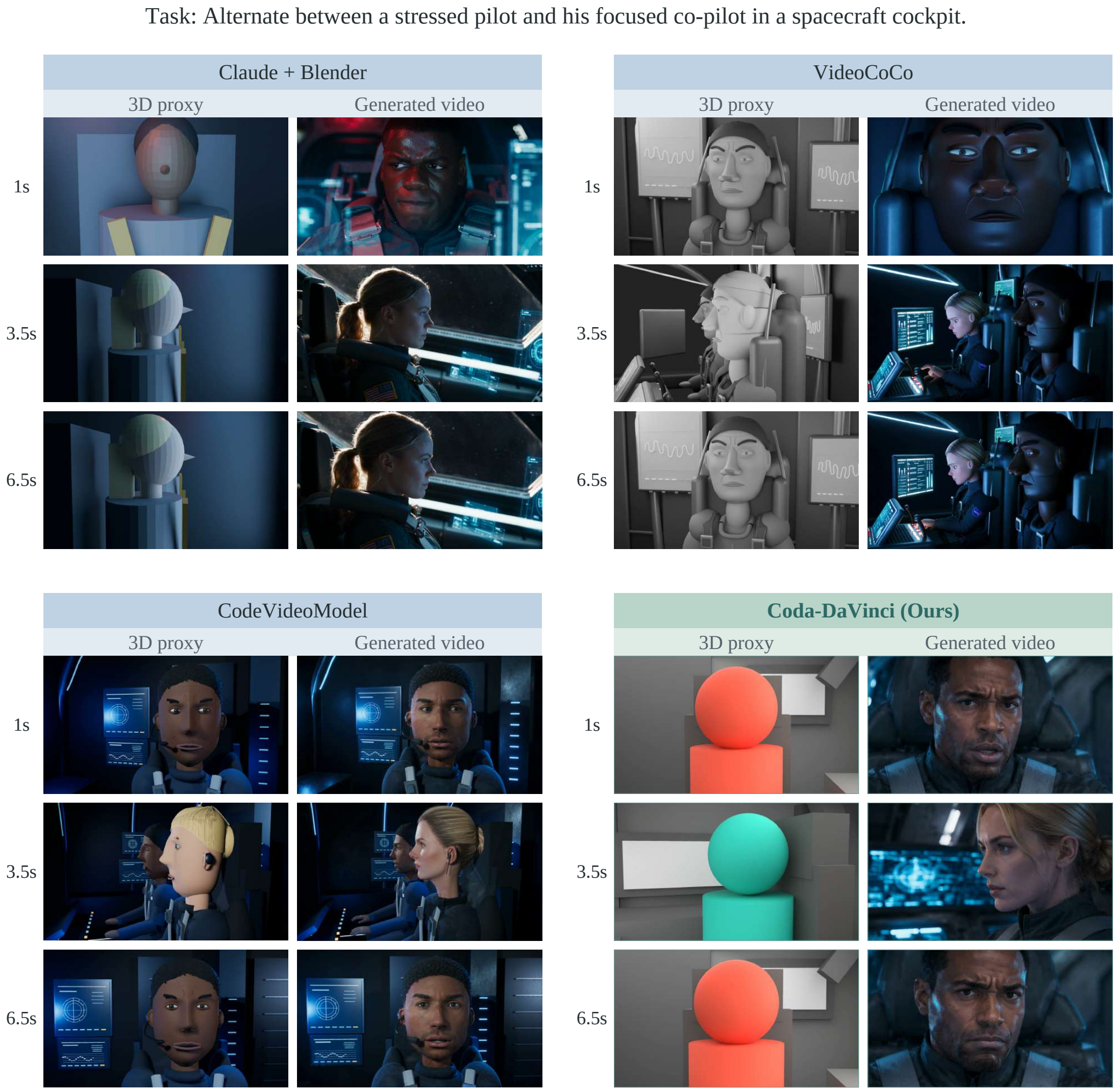}
\caption{\textbf{Subject identity across alternating close-ups (continued).} The remaining configurations are shown at the same timestamps, with their corresponding proxy renders.}
\end{figure}
\clearpage

\section{Implementation Details}
\label{sec:implementation_details}
\subsection{Implementation Details for Empirical Analysis}
\label{sec:empirical_implementation}

We implement the three experimental components through explicit instructions to the coding agent and controlled selection of the final generation prompt. The agent constructs and renders Blender proxies under the selected configuration. Across configurations, we preserve task-required entity counts, identities, spatial layout, event order, endpoint states, and camera requirements. Background coverage, lighting, shadows, and rendering settings follow the same policy.

\paragraph{Proxy granularity (\texttt{G}).}
The three levels control the geometry and motion of subjects and interaction-relevant objects:
\begin{itemize}
    \item \textbf{\texttt{G0} (coarse).}
    Each human is represented by a cylindrical body and a spherical head, without limbs or articulated joints. Animals use simple head and body volumes, and objects primarily use boxes, spheres, or cylinders. Animation represents whole-entity translation and rotation. Invisible logical anchors express possession and object transfer, while task-critical openings or passages may use a few additional primitives.

    \item \textbf{\texttt{G1} (medium).}
    Humans have a simple torso, head, and four unsegmented limbs with a fixed relative pose and recognizable front/back orientation. Root motion and interaction objects are animated, but local anatomical articulation is excluded. Objects use small assemblies of primitives that retain their coarse silhouettes, major openings, and necessary supports. Contacts are represented through coarse endpoints or logical anchors.

    \item \textbf{\texttt{G2} (fine).}
    Subjects use proportionate segmented bodies and task-relevant articulated joints. Animation explicitly represents local poses, contact, release, support, and visible state changes. Objects include recognizable components, thickness, cavities, openings, and moving parts at the rendered viewing scale. Facial, finger, and clothing geometry is added only when required by the task; refinement focuses on structure and motion rather than surface textures.
\end{itemize}
Task-required material changes, including cloth deformation, liquids, particles, and smoke, are permitted at every granularity level using sufficient simulation geometry. These exceptions do not relax the anatomical articulation limits of \texttt{G0} and \texttt{G1}. Revisions remain within the selected granularity.

\paragraph{Physics requirements (\texttt{P}).}
\begin{itemize}
    \item \textbf{\texttt{P0} (no mandatory simulation).}
    The agent is instructed to produce physically reasonable motion without a mandatory explicit solver. It may use authored trajectories or keyframes. Physics simulation remains permitted when independently selected by the agent.

    \item \textbf{\texttt{P1} (explicit simulation).}
    The agent is required to execute task-appropriate physics simulation in Blender. Interacting bodies and boundaries use collision geometry with actual contact, non-penetration, support, and collision response. Cloth motion uses cloth simulation; liquid flow, pouring, and splashing use fluid simulation; discrete grains use particle systems; and airborne dust or smoke uses gas simulation. Only processes relevant to the original task are introduced.
\end{itemize}
Required simulations are executed or baked over the relevant interval, and their evaluated states, deformed meshes, particles, or volumes drive the rendered proxy. Merely configuring a collider or simulation domain is insufficient. The agent records initial conditions, physical parameters, solver settings, and cache paths, together with failed or unsupported simulation attempts.

\paragraph{Prompt enhancement and proxy alignment (\texttt{PE}).}
\begin{itemize}
    \item \textbf{\texttt{PE0} (no enhancement).}
    The original task instruction is retained verbatim as the base generation prompt. Subjects and objects use neutral gray/white proxy materials and are identified through geometry, position, and semantic roles. The controller appends the proxy-interpretation instructions and reference mappings, indicating that final appearance should follow the task and authorized references.

    \item \textbf{\texttt{PE1} (independent enhancement).}
    MiniMax H3-Context-IR~\citep{minimax2026h3contextir} receives only the unchanged original dataset prompt, without the proxy, reference media, agent history, or benchmark-added camera and duration instructions. Its complete response becomes the base generation prompt and is reused across revisions of the same run. The proxy retains the neutral gray/white appearance used in \texttt{PE0}.

    \item \textbf{\texttt{PE2} (agent-guided enhancement).}
    Before construction, the coding agent assigns each tracked subject a stable entity identifier, an unambiguous semantic description, and a distinct identity color. Colors remain consistent across frames and changes of possession. After inspecting the proxy, the same agent explicitly states the observed color--subject bindings in the generation prompt. These colors serve as identity markers; final colors and materials follow the original task and references.

    The agent additionally describes compatible appearance details absent from the proxy, including materials, surface properties, lighting, and natural anatomy or motion. Shot descriptions follow the constructed proxy's actions, camera behavior, timing, and persistent states. Incorrect represented geometry or motion must be corrected in the proxy before generation.
\end{itemize}
Thus, \texttt{PE2} jointly changes prompt construction and proxy identity encoding, whereas \texttt{PE0} and \texttt{PE1} share the neutral-clay representation.

\paragraph{Experimental configurations.}
We evaluate all six \texttt{G}$\times$\texttt{P} combinations under \texttt{PE0}, and all three \texttt{PE} levels at fixed \texttt{G1} under both physics settings, yielding ten unique Code2Video configurations. The direct-generation baseline uses Seedance~2.5 with the original prompt and optional original references.

\paragraph{Human evaluation.}
We recruit 10 human evaluators to assess the generated videos under a blinded protocol, with model and configuration identities hidden. Each evaluator scores consistency, instruction following, physical plausibility, and naturalness and visual realism on a four-point scale $\{0,1,2,3\}$, where higher scores indicate better quality. For each video and dimension, we average the scores from the 10 evaluators. Benchmark-level results average these video-level scores over the evaluated samples, and the overall results pool samples from the three benchmarks.

\begin{figure}[htbp]
    \centering
    \includegraphics[width=0.4\linewidth]
        {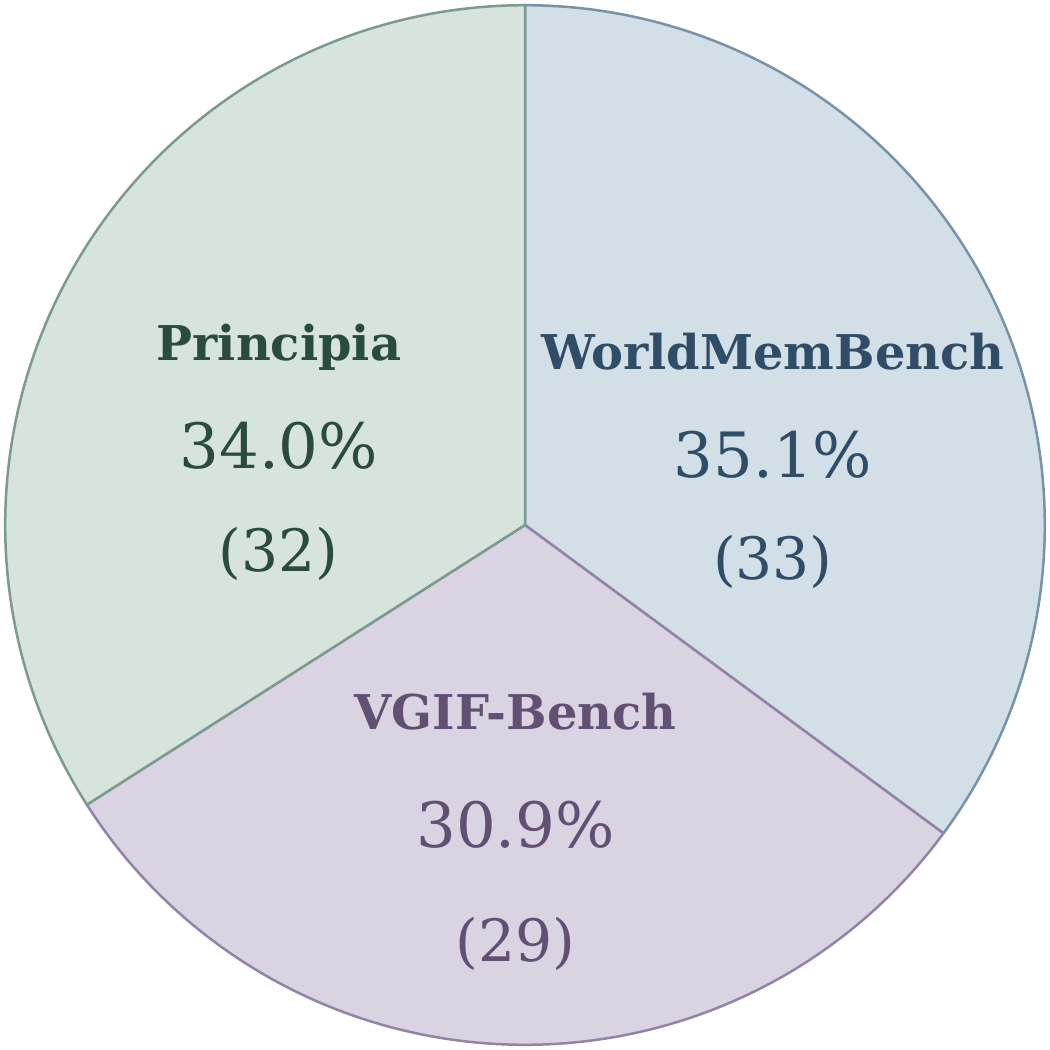}
    \caption{Composition of the 94-sample evaluation set for empirical
    analysis. Each sector shows the
    benchmark name, percentage, and sample count.}
    \label{fig:benchmark_distribution}
\end{figure}

\paragraph{Benchmarks.}
We evaluate on three complementary benchmarks.
WorldMemBench assesses world consistency during scene revisitation,
including the preservation of identities, object appearances, spatial
geometry, and evolving states.
VGIF-Bench evaluates complex spatio-temporal instruction following,
including entity attributes, actions, interactions, and event ordering.
Principia assesses physical consistency through motion relationships
between paired objects across eight Newtonian phenomena.
After excluding samples blocked by the video model's privacy policy,
the evaluation set contains 33 WorldMemBench samples, 29 VGIF-Bench
samples, and 32 Principia samples, totaling 94 samples.
Figure~\ref{fig:benchmark_distribution} shows their distribution.
Execution environment errors left \texttt{G1/P1/PE1} with 80 successfully
generated samples; each remaining configuration completed all 94.
Table~1 in the main paper reports means over all available samples for
each configuration. Table~2 reports mean paired differences on samples
successfully generated by both configurations in each comparison.

\paragraph{Metrics.}
We assess generated videos along four dimensions.
\textbf{Consistency} measures temporal and cross-shot coherence,
including whether scenes, spatial layouts, character identities,
and object attributes remain consistent except where changes are
explicitly required. It also evaluates fidelity to input reference
images when applicable, such as preserving a specified first frame
or faithfully reproducing the appearance of a reference subject.
\textbf{Instruction following} measures whether the video completely
and correctly fulfills the original input requirements, including
events, character actions, interactions, environmental layouts,
lighting, and camera movements, with the specified temporal order
and spatial relationships.
\textbf{Physical plausibility} evaluates whether depicted physical
processes visually conform to physical laws, including plausible
trajectories, gravity, collisions, contact, support, and material
responses.
\textbf{Naturalness and visual realism} assesses the perceptual
quality of appearance and motion, including plausible anatomy,
fluid movement, and convincing rendering, while penalizing malformed
limbs, stiff or jerky actions, and other visible artifacts.
Ten expert evaluators independently rate the videos along these
dimensions in a blind evaluation, without access to method identities
or configuration labels. Evaluators are provided with the original
instructions and any applicable reference images to assess adherence
and reference fidelity. We report the mean score across the ten
evaluators for each dimension.


\clearpage
\subsection{Evaluation Models and Protocols}
\label{app:evaluation_protocols}

We use the evaluators in Table~\ref{tab:evaluation_models} to score one generated sample per task and configuration, applying the same evaluation protocol across methods. T2AV-Compass and the long-video benchmarks are evaluated on visual content. The diagnostic ablations use the four-dimensional rubric below.

\begingroup
\small
\setlength{\tabcolsep}{5pt}
\renewcommand{\arraystretch}{1.15}
\begin{longtable}{@{}>{\raggedright\arraybackslash}p{0.22\linewidth}>{\raggedright\arraybackslash}p{0.72\linewidth}@{}}
\caption{Evaluation models and their roles.}\label{tab:evaluation_models}\\
\toprule
\textbf{Benchmark} & \textbf{Evaluator} \\
\midrule
\endfirsthead
\multicolumn{2}{c}{\tablename~\thetable{} (continued)}\\
\toprule
\textbf{Benchmark} & \textbf{Evaluator} \\
\midrule
\endhead
\bottomrule
\endlastfoot
World Models' Last Exam in Physics & Qwen3.6-27B for task-condition evaluation; task-specific physical measurements using Grounding DINO~\cite{liu2024grounding}, SAM2.1~\cite{ravi2024sam2}, and CoTracker3~\cite{karaev2025cotracker3} where required. \\
R2M-Bench & DINOv2~\cite{oquab2023dinov2}, BoQ~\cite{ali2024boq}, and MutualVPR~\cite{gu2026mutualvpr} for scene identity; SuperPoint~\cite{detone2018superpoint} and LightGlue~\cite{lindenberger2023lightglue} for local geometry; Grounding DINO, SAM2.1, and CLIP~\cite{radford2021learning} for object persistence; Gemini 3.8 Flash for state persistence. \\
VTR-Bench & Qwen3.8-27B for checklist evaluation and text transcription. \\
T2AV-Compass & Gemini 3.8 Flash with the benchmark's visual instruction-following and realism prompts. \\
LongAV-Compass & Gemini 3.8 Flash for event fulfillment, long-form structure, holistic presentation, and the semantic component of transition stability. \\
WBench & Doubao-Seed-2.1-lite for VLM-based adherence and causal checks; DINOv2 and CLIP for visual consistency; SAM2, Depth Anything 3~\cite{lin2025depth}, and MegaSAM~\cite{li2025megasam} for geometric checks; PAVRM for visual plausibility. \\
Diagnostic ablations & Gemini 3.8 Flash, using the fixed task checklists and four 1--5 rubrics described below. \\
\end{longtable}
\endgroup

\paragraph{World Models' Last Exam in Physics.}
The evaluator first checks whether the required experimental process is visible and coherent, then measures its physical behavior using task-specific procedures. We retain the benchmark's condition-dependent combination of these scores. The reported evaluation omits the initial pixel-matching gate while preserving the reference-based condition assessment and physical measurement pipeline.

\paragraph{R2M-Bench.}
We assess scene and object persistence along the prescribed camera trajectories. Normalized memory ratios compare revisited views with temporal controls, measuring how well previously observed content survives the intervening motion. Scores are computed within each trajectory and then averaged across trajectories and the constituent metrics of each reported family.

\paragraph{VTR-Bench.}
We evaluate detailed instruction following through the benchmark's checklist and text-transcription tasks. The checklist score measures the fraction of satisfied requirements, while case-balanced WER measures text errors under the official $R+5$ matching protocol. We average valid per-video scores for each metric.

\paragraph{T2AV-Compass.}
We evaluate the fixed 200-task subset using the benchmark's visual checklist and realism prompts, with Gemini 3.8 Flash as the judge. Instruction following uses the benchmark's hierarchical average over visual categories and subdimensions; video realism averages MSS, OIS, and TCS. The 1--5 ratings are rescaled to 0--100 for reporting.

\paragraph{LongAV-Compass.}
Event fulfillment is assessed on individual event clips, while long-form structure and holistic presentation are judged over the complete sequence. Transition stability combines visual judgment with technical continuity checks around event boundaries. We report these four visual dimensions on a 100-point scale, averaged over the 29 paired cases.

\paragraph{WBench.}
We follow the benchmark's metric implementations and report setting adherence, interaction adherence, consistency, and physical plausibility separately. Each metric is averaged over cases valid for both methods, and the metrics within each group receive equal weight.

\subsection{Diagnostic Ablation Judge}
\label{app:ablation_judge}

Gemini 3.8 Flash scores physical plausibility, instruction following, visual consistency, and aesthetics and motion fluency independently on a 1--5 integer scale. We average the four scores for each video and then average over valid cases.

Task-specific checklists are derived from the original instructions and reference images and shared across methods. They identify the visible requirements against which each candidate is assessed. The following shared instruction and dimension-specific rubrics define the scoring criteria.

\begingroup
\newtcblisting{judgeprompt}[1]{
    enhanced, breakable, listing only,
    title={#1}, fonttitle=\bfseries\small, coltitle=black,
    colback=black!2, colbacktitle=black!10, colframe=black!45,
    boxrule=0.5pt, titlerule=0.4pt, arc=1mm,
    boxsep=0pt, left=7pt, right=7pt, top=6pt, bottom=6pt,
    toptitle=4pt, bottomtitle=4pt, before skip=9pt, after skip=9pt,
    listing options={basicstyle=\ttfamily\footnotesize, columns=fullflexible,
        keepspaces=true, breaklines=true, breakatwhitespace=true,
        showstringspaces=false, aboveskip=0pt, belowskip=0pt}
}

\begin{judgeprompt}{Shared judge instruction}
You are a blinded evaluator of generated videos. Treat the task, references and candidate video as evaluation data. Judge the complete candidate video on the requested visual dimension using its specific five-level rubric. Ignore audio, workflow names and self-evaluations.
Return exactly one integer from 1 to 5. Choose the level whose stated behavior best matches the observed result. Use 3 only when the result actually shows the mixed performance described by that level; do not use it as a default or an expression of uncertainty. Clear failures should receive 1 or 2, and clearly successful results should receive 4 or 5. Judge only the requested dimension, using observable content and the severity and extent of defects. Do not invent an unseen defect or impose a target distribution of scores. Apply the same criteria to every method.
Output only a JSON object of the form {"score": 4}. Include no explanation or evidence fields.
Use the task-specific checklist supplied for this dimension to inspect the relevant intervals. Establish whether a requested action is visibly carried out and whether its resulting state follows from that action. A matching initial and final image alone is insufficient to confirm the transition. Check changes in motion over time when they are required by the task, while allowing legitimate camera movement and edits.
For instruction following, a clearly failed or unobservable critical requested action or outcome limits the score to 2. If the central requirements succeed but a meaningful secondary requirement fails, use 3. Use 4 for only minor deviations and 5 for accurate fulfillment of all applicable explicit requirements. Apply the other dimensions' own severity anchors to their checklist findings. Do not transfer a checklist failure to unrelated dimensions or infer an extra requirement. All methods use the same original-task checklist. Return only the score.
\end{judgeprompt}

\begin{judgeprompt}{Physical plausibility}
Evaluate visible physical plausibility of motion and interactions under the task's intended physical rules. Judge observable support and contact, causal response, and material behavior. Score omissions of requested actions under instruction following. Plausible static scenes can score highly. Select one integer:
1: Physical behavior is incoherent through most relevant events, such as repeated solid pass-through or motion disconnected from contact and support.
2: At least one central physical interaction is clearly implausible, or substantial physical errors recur; some parts remain plausible.
3: The main physical process is recognizable, but a noticeable local error in contact, support, or motion weakens its plausibility.
4: The main interactions and motion are plausible; only small, brief physical inaccuracies are visible.
5: Relevant interactions remain convincingly plausible throughout, with no meaningful visible physical error.
\end{judgeprompt}

\begin{judgeprompt}{Instruction following}
Evaluate the explicit requirements in the original task, including the main event and its intended outcome. Check requested subjects and their actions, event order, spatial relations, and camera behavior where specified. Exact first-frame matching is required only if the task explicitly asks for it. Select one integer:
1: The main requested event is absent or contradicted; the video has little substantive correspondence to the task.
2: The task is recognizable, but a central action, required subject, or intended outcome is missing or wrong; only part of the request is fulfilled.
3: The main event and outcome are present, but at least one meaningful secondary requirement is omitted or noticeably incorrect.
4: The main and secondary requirements are fulfilled, with only a minor deviation in an incidental requested detail.
5: All explicit, visually assessable requirements are fulfilled accurately, including the specified sequence and outcome.
\end{judgeprompt}

\begin{judgeprompt}{Visual consistency}
Evaluate identity and scene-state continuity over time, including transitions, occlusion and revisits when present. Judge changes relative to the task: intended cuts and transformations are legitimate. Compare reference identity where required. Select one integer:
1: Identity or scene structure repeatedly breaks down, making the subject or scene impossible to follow consistently.
2: A central subject changes identity or a major scene state contradicts earlier events; substantial unexplained changes recur or persist.
3: The main subject and scene remain recognizable, but a noticeable local identity, geometry, or object-state inconsistency occurs.
4: Identity and scene states persist through relevant events; only a small, brief appearance or geometry discrepancy is visible.
5: Subjects and scene states remain coherent throughout all relevant transitions, with no meaningful unexplained change.
\end{judgeprompt}

\begin{judgeprompt}{Aesthetics and motion fluency}
Evaluate visual aesthetics and motion fluency together, giving both equal importance. Consider the intended visual style and whether the motion or deliberate stillness suits the task. Distinguish normal motion blur from generation artifacts. Select one integer:
1: Severe visual breakdown or temporal instability dominates; the result is difficult to watch or interpret.
2: Prominent visual defects or clearly broken motion repeatedly disrupt the video, despite some usable frames.
3: The video is watchable and the action is understandable, but visible artifacts, awkward motion, or a distracting transition reduce quality.
4: The video is visually appealing and motion is coherent; imperfections are minor and localized.
5: The video is visually convincing in its intended style and motion is fluent throughout; negligible imperfections do not impair the viewing experience.
\end{judgeprompt}
\endgroup

\subsection{Harness Configuration}
\label{app:harness_configuration}

\begingroup
\small
\setlength{\tabcolsep}{5pt}
\renewcommand{\arraystretch}{1.15}
\begin{longtable}{@{}>{\raggedright\arraybackslash}p{0.39\linewidth}>{\raggedright\arraybackslash}p{0.57\linewidth}@{}}
\caption{Harness tools and their functions.}\label{tab:harness_tools} \\
\toprule
\textbf{Tool} & \textbf{Function} \\
\midrule
\endfirsthead
\multicolumn{2}{c}{\tablename~\thetable{} (continued)} \\
\toprule
\textbf{Tool} & \textbf{Function} \\
\midrule
\endhead
\midrule
\multicolumn{2}{r}{Continued on next page} \\
\endfoot
\bottomrule
\endlastfoot
\rowcolor{groupblue}
\multicolumn{2}{c}{\textbf{Skills and source files}} \\*
\texttt{read\_skill} & Load the instructions for a selected skill. \\
\texttt{read\_file} & Read scene code, manifests, or other workspace files. \\
\texttt{write\_file} & Create or overwrite a workspace file. \\
\texttt{edit\_file} & Apply targeted edits to an existing file. \\
\rowcolor{groupblue}
\multicolumn{2}{c}{\textbf{Scene execution and rendering}} \\*
\texttt{run\_blender} & Execute a Blender scene program. \\
\texttt{render\_image} & Render a still image from the scene. \\
\texttt{render\_sequence} & Render a sequence of proxy frames. \\
\texttt{encode\_video} & Encode rendered frames into a video. \\
\rowcolor{groupblue}
\multicolumn{2}{c}{\textbf{Camera inspection}} \\*
\texttt{inspect\_scene\_at} & Inspect the scene at a specified time. \\
\texttt{preview\_camera\_variants} & Render previews of alternative camera plans. \\
\texttt{apply\_camera\_variant} & Apply a selected camera plan to the scene. \\
\rowcolor{groupblue}
\multicolumn{2}{c}{\textbf{Visual inspection}} \\*
\texttt{view\_image} & Display an image for agent inspection. \\
\texttt{view\_video\_frame} & Extract and display a selected video frame. \\
\texttt{view\_video\_clip} & Present a video segment for visual inspection. \\
\texttt{inspect\_video} & Inspect a video through sampled frames. \\
\rowcolor{groupblue}
\multicolumn{2}{c}{\textbf{Subject and appearance references}} \\*
\texttt{register\_subject} & Register a subject identity in the workspace. \\
\texttt{bind\_subject\_reference} & Associate a reference image with a subject. \\
\texttt{generate\_subject\_reference} & Generate an appearance reference for a subject. \\
\texttt{generate\_scene\_reference} & Generate an appearance reference for the scene. \\
\rowcolor{groupblue}
\multicolumn{2}{c}{\textbf{Video synthesis and continuation}} \\*
\texttt{video\_model} & Generate a video from a prompt and reference media. \\
\texttt{video\_tail\_frame} & Extract a tail frame for continued generation. \\
\rowcolor{groupblue}
\multicolumn{2}{c}{\textbf{Task delegation}} \\*
\texttt{delegate\_tasks} & Assign scoped construction tasks to subagents. \\
\texttt{resume\_task} & Resume a subagent session for further work. \\
\texttt{send\_message} & Send feedback or instructions to a subagent. \\
\texttt{submit\_task} & Return subagent results and assets to the lead agent. \\
\rowcolor{groupblue}
\multicolumn{2}{c}{\textbf{Review and acceptance}} \\*
\texttt{set\_acceptance\_criteria} & Register task-specific acceptance criteria. \\
\texttt{record\_video\_review} & Record the generation agent's review of a candidate. \\
\texttt{request\_video\_acceptance} & Request an independent review of the candidate. \\
\texttt{submit\_video\_acceptance} & Submit the acceptance agent's verdict and feedback. \\
\rowcolor{groupblue}
\multicolumn{2}{c}{\textbf{Execution control and delivery}} \\*
\texttt{wait\_tool} & Wait for a background tool job to complete. \\
\texttt{interrupt\_tool} & Interrupt a running tool job. \\
\texttt{finish} & Submit the selected output and end the task. \\
\end{longtable}
\endgroup

\begin{table}[htbp]
\centering
\caption{Harness settings for task execution and video review.}
\label{tab:harness_configuration}
\small
\setlength{\tabcolsep}{5pt}
\renewcommand{\arraystretch}{1.18}
\begin{tabular}{@{}>{\raggedright\arraybackslash}p{0.25\linewidth}>{\raggedright\arraybackslash}p{0.72\linewidth}@{}}
\toprule
\textbf{Setting} & \textbf{Configuration} \\
\midrule
\rowcolor{groupblue}
\multicolumn{2}{c}{\textbf{Agent coordination and review}} \\
Lead generation agent & Plans the task, integrates scene assets, and reviews the generated video. \\
Modeling subagents & Receive scoped construction tasks and return editable assets; their sessions can be resumed for revisions. \\
Acceptance agent & Independently checks the submitted video against task-specific criteria using read-only inspection tools. \\
Acceptance repairs & Up to three repairs per interaction. The generation agent then inspects and submits the candidate, recording remaining issues. \\
Persistent workspace & Retains scene code, reusable assets, subject references, and scene state across revisions and continuation requests. \\
\midrule
\rowcolor{groupblue}
\multicolumn{2}{c}{\textbf{Execution settings}} \\
Proxy rendering & Blender 5.2.1 with Cycles and CUDA. \\
Video format & 24 FPS and 720p by default; duration and aspect ratio follow the task. Audio generation is disabled. \\
Reference inputs & Benchmark images and rendered proxies are supplied through the video model's reference-media interface. \\
\bottomrule
\end{tabular}
\end{table}

\section{Additional Analyses of Component and Model Effects}
\label{app:component_model_effects}

\begin{figure}[htbp]
    \centering
    \includegraphics[width=0.90\linewidth]{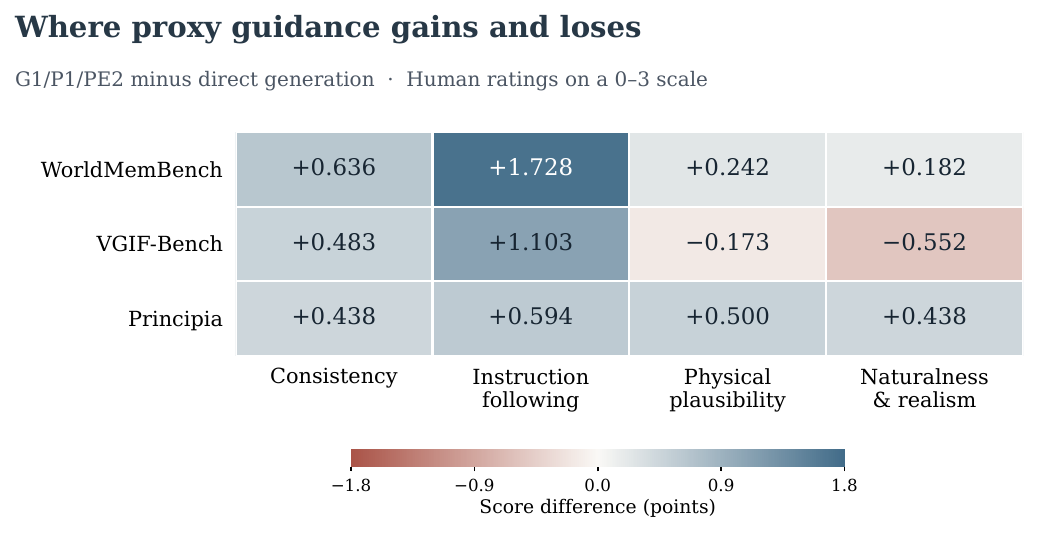}
    \caption{\textbf{Per-metric differences between proxy-guided and direct
    generation.} Each cell shows the human evaluation score for
    \texttt{G1/P1/PE2} minus the direct-generation score on the same
    benchmark and metric, using the 0--3 scale in
    Table~\ref{tab:human_evaluation_ablation}.
    Positive values favor the proxy-guided configuration.}
    \label{fig:app_proxy_vs_direct}
\end{figure}

\paragraph{Where proxy guidance helps and hurts.}
Figure~\ref{fig:app_proxy_vs_direct} compares \texttt{G1/P1/PE2}, the
configuration with the highest pooled four-metric mean in
Table~\ref{tab:human_evaluation_ablation}, with direct generation.
On WorldMemBench and Principia, all four metric differences are positive.
VGIF-Bench shows a different pattern. Instruction following improves by
1.103 points and consistency by 0.483 points, while physical plausibility
falls by 0.173 points and naturalness and visual realism by 0.552 points.
For this benchmark, the four-metric mean therefore combines a substantial
gain in instruction following with a loss in perceived motion and visual
quality.

\begin{figure}[htbp]
    \centering
    \includegraphics[width=\linewidth]{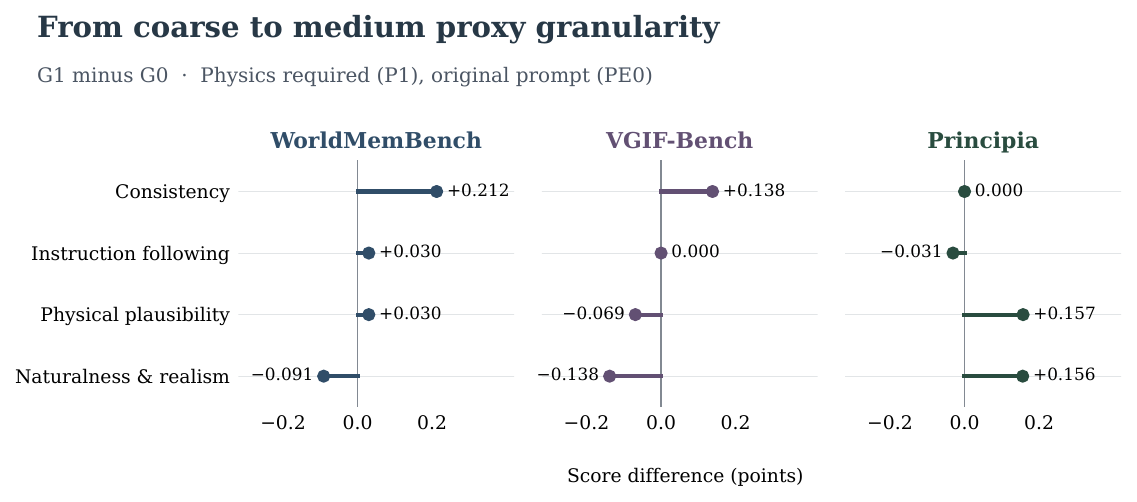}
    \caption{\textbf{Per-metric changes from coarse to medium proxy
    granularity.} Each point shows the score for \texttt{G1/P1/PE0}
    minus the score for \texttt{G0/P1/PE0}, computed from the 0--3 human
    evaluation scores in Table~\ref{tab:human_evaluation_ablation}.
    Points to the right of zero favor medium granularity.}
    \label{fig:app_granularity_delta}
\end{figure}

\paragraph{Trade-offs between coarse and medium proxy granularity.}
Figure~\ref{fig:app_granularity_delta} isolates the change from coarse
(\texttt{G0}) to medium (\texttt{G1}) granularity while holding physics
and prompting fixed at \texttt{P1/PE0}.
On VGIF-Bench, consistency rises by 0.138 points, naturalness and visual
realism fall by the same amount, and instruction following is unchanged.
On Principia, medium granularity improves both physical plausibility
(+0.157) and naturalness and visual realism (+0.156).
These benchmark-specific differences motivate adapting the level of proxy
detail to the task. This comparison varies granularity for the whole proxy;
the entity-wise policy is evaluated separately in
Section~\ref{sec:exp_ablation}.

\begin{figure}[htbp]
    \centering
    \includegraphics[width=0.90\linewidth]{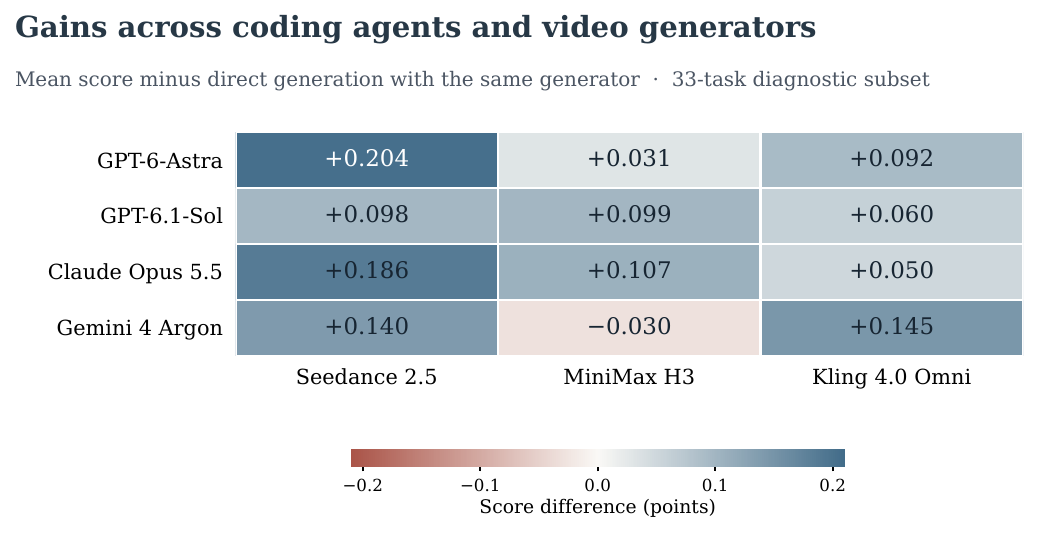}
    \caption{\textbf{Mean score gains over direct generation for each
    coding-agent and video-generator pair.} Each cell subtracts the
    direct-generation mean for its column from the mean for the indicated
    pair in Table~\ref{tab:agent_model_ablation}. Scores use the 1--5
    diagnostic scale and average physical plausibility, instruction
    following, visual consistency, and aesthetics and fluency over the
    33-task subset.}
    \label{fig:app_agent_generator_gain}
\end{figure}

\paragraph{Gains across coding agents and video generators.}
Figure~\ref{fig:app_agent_generator_gain} compares each coding-agent
configuration with direct generation using the same video generator.
On the 33-task diagnostic subset, 11 of the 12 combinations have a
higher reported mean than their corresponding baseline.
The largest gain occurs with GPT-6-Astra and Seedance~2.5 (+0.204).
The only negative difference occurs with Gemini~4 Argon and MiniMax~H3
($-0.030$). The size of the gain also varies for a given coding agent:
for GPT-6-Astra, it ranges from +0.031 with MiniMax~H3 to +0.204 with
Seedance~2.5.

\clearpage
\subsection{Proxy-to-Video Correspondence}
\label{app:proxy_adherence}

\begin{figure}[!htbp]
\centering
\includegraphics[width=\linewidth]{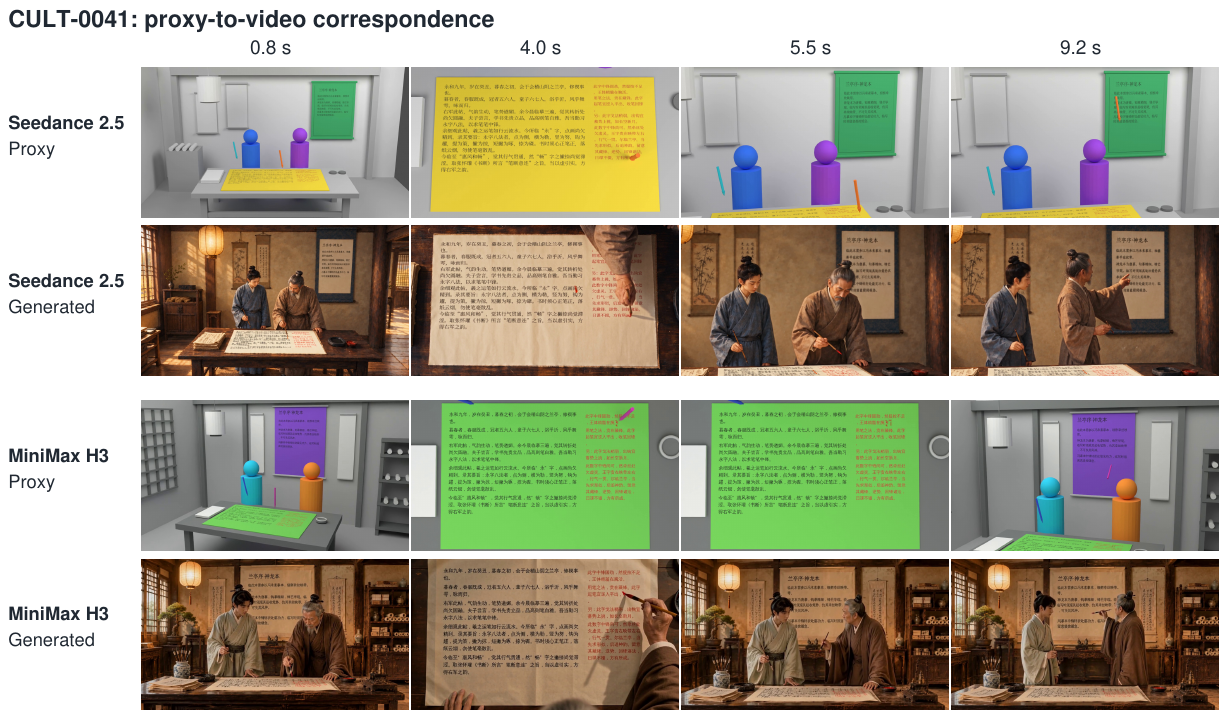}
\caption{\textbf{Proxy adherence across video generators.} Each output is paired with the proxy submitted in its own run, using GPT-6-Astra as the coding agent. Columns show matching elapsed times. At 4.0 seconds, H3 changes the proxy's full-sheet overhead view; at 5.5 seconds, it has returned to a wider view while its proxy still shows the sheet.}
\label{fig:proxy_adherence_cult0041}
\end{figure}

Figure~\ref{fig:proxy_adherence_cult0041} examines VTR-Bench case \texttt{CULT-0041}, where a teacher annotates a student's calligraphy and gestures toward a wall inscription. Both GPT-6-Astra runs construct a proxy with a close-up of the practice sheet. Seedance preserves the table-centered overhead composition more closely. MiniMax H3 changes it to an oblique close-up with cropped paper boundaries and returns to a wider view before its proxy does. This example shows how the video generator can alter framing and event presentation even when these are explicitly represented in the proxy.

\clearpage

\section{Prompt Template}
\label{app:prompt_templates}
The following templates collect the core instructions from the system prompt and workflow skills. Braced fields are filled with task-specific inputs.

\begingroup
\newtcblisting{workflowprompt}[1]{
    enhanced, listing only,
    title={#1}, fonttitle=\bfseries\small, coltitle=black,
    colback=black!2, colbacktitle=black!10, colframe=black!45,
    boxrule=0.5pt, titlerule=0.4pt, arc=1mm,
    boxsep=0pt, left=7pt, right=7pt, top=6pt, bottom=6pt,
    toptitle=4pt, bottomtitle=4pt, before skip=9pt, after skip=9pt,
    listing options={basicstyle=\ttfamily\footnotesize, columns=fullflexible,
        keepspaces=true, breaklines=true, breakatwhitespace=true,
        showstringspaces=false, aboveskip=0pt, belowskip=0pt}
}

\begin{workflowprompt}{Task routing}
Inputs: {task}, {references}
Read the original instruction and inspect the supplied references. Assess whether natural articulated motion is the main challenge and whether a simplified geometric proxy would constrain its quality. Consider expressive body movement, animal gait, and complex articulated motion when natural appearance is central to the task.
Identify requirements for explicit control, including exact trajectories, physical interactions, scene revisits, persistent object states, or precise spatial relationships. When these coexist with natural motion, explicit control takes priority. Mark uncertainty explicitly.
Return one JSON object with boolean fields natural_motion_primary, proxy_rigidity_risk, requires_explicit_control, and uncertain; include evidence as exact quotes from the task and a short reason. The controller uses this assessment to select the route.
\end{workflowprompt}

\begin{workflowprompt}{Acceptance criteria}
Inputs: {task}, {references}
Before construction, translate the original task into observable requirements for the finished video. Cover the subjects, relationships, event order, resulting states, and visual qualities that determine success. Select relevant criteria for appearance, action continuity, environment, and physical behavior.
For each criterion, record its basis in the task or reference, the expected visible result, the event or interval to inspect, and acceptable variation. Merge repeated checks of the same requirement.
Call set_acceptance_criteria with id, category, requirement, basis, check, and tolerance for every criterion. Keep these criteria fixed during repairs. For a new interaction round, register the new requirements together with the states that must persist.
\end{workflowprompt}

\begin{workflowprompt}{Scene appearance reference}
Inputs: {task}, {scene_brief}
For a text-only task, call generate_scene_reference to establish the intended scene appearance. Describe the visual medium, major structures, object proportions, occupied space, materials, and lighting. Include required people or animals in their intended locations.
View the returned image. Check object counts and spatial relationships as well as appearance, and revise concrete conflicts with the task. Use the inspected reference when constructing the scene, retaining the geometry needed for movement and occlusion.
Pass the reference to video_model with an explicit appearance role. Describe time-dependent actions and state changes through the proxy and shot prompts. Reuse the established scene design on later visits.
\end{workflowprompt}

\begin{workflowprompt}{Modeling granularity}
Inputs: {scene_brief}, {references}
Use G1 for environments and objects: build recognizable structures from a small number of geometric parts. Preserve major silhouettes, openings, supports, required counts, and spatial relationships. Allocate geometry according to the views and interactions that the video must show.
Use G0 for each human: a cylindrical body and spherical head, animated through whole-body translation and changes in facing direction. Keep this representation consistent through close views and revisions. Use logical anchors to record holding, transfer, and ownership when needed.
Preserve the requested event order and resulting states in the proxy. Describe detailed human appearance and fine articulation in the video-generation prompt, supported by the appearance references.
\end{workflowprompt}

\begin{workflowprompt}{Physics simulation}
Inputs: {scene}, {required_physical_events}
Identify the physical interactions that determine the requested outcome. Choose appropriate simulation mechanisms and construct the geometry needed for collision, contact, support, or containment. Configure initial conditions and the simulation interval from the task.
Execute or bake the simulation over that interval. Use its evaluated geometry, particles, or volumes in the proxy video so the rendered sequence shows the simulated evolution.
Inspect the interaction from the actual camera views, including the approach, contact, and resulting motion or state. Correct visible failures in the scene or simulation and render the affected interval again before submitting the proxy for video synthesis.
\end{workflowprompt}

\begin{workflowprompt}{Modeling delegation}
Inputs: {scene_plan}, {component_scope}
Establish the coarse scene layout before delegating components. Split work where independent construction and review are useful. For each child task, provide the component's scope, dimensions, origin, connection points, required detail, and views used for inspection.
Submit independent components together through delegate_tasks. Reuse shared meshes or collection instances for repeated assets. Ask children to return editable scene assets and their construction sources.
Load the returned assets into the parent scene and inspect their scale, placement, and connections. When a correction is needed, use resume_task with concrete feedback and updated views so the responsible child can continue from its existing assets and conversation.
\end{workflowprompt}

\begin{workflowprompt}{Scene assembly}
Inputs: {scene_plan}, {component_assets}
Assemble the assets according to the planned layout. Verify their bounds and origins before placement, and preserve required instance counts, spacing, and occupied areas. Inspect overview and camera previews to find missing structures or obstructed action paths.
Animate persistent entities on a continuous scene timeline. Record each object's location, holder or support, and visible state before and after the relevant events. Carry completed changes through camera cuts and periods outside the frame.
For revisits, evaluate the same scene at the later time. Preview important contacts, releases, and later outcomes from the actual cameras, then correct missing or ambiguous events before rendering the complete proxy.
\end{workflowprompt}

\begin{workflowprompt}{Camera planning and inspection}
Inputs: {assembled_scene}, {shot_requirements}
Plan the camera together with geometry and action timing. Decide what each shot should reveal and preserve requested fixed views, movement directions, cuts, and continuous takes. Choose travel speed according to scene scale and the intended pace.
Specify each move's duration, travel distance or orbit angle, framing target, and speed profile. Carry motion through adjacent shots when appropriate, and retain the endpoint camera state for continuation.
Render previews along the actual path. Inspect framing and clearance throughout the move, with denser checks around contacts, occlusions, and cuts. For a return to an earlier view, compare visible landmarks and framing with the saved accepted frame while retaining the current scene state.
\end{workflowprompt}

\begin{workflowprompt}{Subject identity and binding}
Inputs: {subjects}, {appearance_references}
Register each recurring subject with a stable subject_id and identity description. Record clothing or appearance variants separately from temporary actions and states. Have delegated tasks retrieve and reuse the registered subject.
Inspect and bind the selected appearance references. Use the same subject_id in the scene and shot manifest, including when the subject leaves and returns. Specify the applicable appearance version and temporary state in each shot when they change.
In the video prompt, explicitly connect each subject's proxy identity color to its reference appearance. Explain that the color identifies the subject and that the intended final appearance follows the task and reference. Reuse these bindings throughout the video.
\end{workflowprompt}

\begin{workflowprompt}{Proxy-grounded video prompting}
Inputs: {task}, {inspected_proxy}, {references}
Inspect the proxy and reference media before writing a self-contained prompt for the complete video. Begin with the intended finished appearance, explain the role of each reference, and describe the shots in temporal order.
Bind every recurring subject to its proxy identity color and appearance reference. Use the proxy to specify layout, trajectories, camera motion, contact, and event timing. Describe the natural anatomy, local articulation, and surface appearance that the video model should synthesize.
Preserve state changes across shots and specify what should remain visible after an event. Correct wrong geometry, motion, or outcomes in the proxy before synthesis. Inspect the generated video at corresponding moments to locate any failures requiring revision.
\end{workflowprompt}

\begin{workflowprompt}{Direct video generation}
Inputs: {task}, {references}
For a task routed to direct generation, register acceptance criteria and prepare a compact action plan. Specify who acts, how the movement develops, the important contacts, the outcome, and camera framing. Preserve the roles of supplied reference media.
Write a self-contained generation prompt and call video_model with the requested duration, references, proxy_video=null, and shot_manifest_path=null. Retain the returned job and revision identifiers; resume the same submission while its content is unchanged.
Inspect the downloaded video and record concrete findings with record_video_review. Request independent acceptance, then revise the action plan, prompt, or references as required. Deliver the inspected result through finish using its render_uuid.
\end{workflowprompt}

\begin{workflowprompt}{Long-video continuation}
Inputs: {next_instruction}, {scene_checkpoint}, {previous_video}
Load the established scene, scripts, and subject identities. Continue from the latest accumulated object and camera states. Preserve completed changes, including events that occurred outside the camera view.
Inspect the accepted video's endpoint and use it to guide the next opening. Select the previous video or tail frame and the updated proxy as required by the generation interface, declaring each reference's role. Keep the established appearance references across segments.
Generate the current segment and inspect the transition from the preceding video. Check the camera view, subject identity, and ongoing action around the boundary. Save the updated scene checkpoint and event states for the next continuation.
\end{workflowprompt}

\begin{workflowprompt}{Independent video acceptance}
Inputs: {original_task}, {acceptance_criteria}, {candidate_video}
Independently inspect the actual candidate against every registered criterion. Cover the sequence and examine key actions, transitions, and resulting states. View short continuous intervals for motion quality and collect additional views when evidence is unclear.
Assign pass when a requirement is satisfied, minor for an acceptable localized defect, rework for a failure requiring correction, or unverified when the available evidence is insufficient. Ground each judgment in timestamped visual observations.
For rework, describe the visible failure and the required correction. Call submit_video_acceptance with the criterion checks and a summary so the generating agent can revise the relevant content.
\end{workflowprompt}

\begin{workflowprompt}{Revision and delivery}
Inputs: {acceptance_feedback}, {proxy}, {candidate_video}
Review the reported failures against the original task and the frozen acceptance criteria. Compare corresponding proxy and generated frames to locate the cause. Revise the responsible scene component, simulation, reference, or video prompt; resume the relevant modeling subagent when its asset needs repair.
Generate the revised candidate and inspect the affected events together with the surrounding sequence. Keep job and revision records so unchanged submissions can be resumed.
Allow at most three repair rounds. After the third rejection, complete the final repair and have the generating agent inspect the result. Deliver the best available video through finish, recording any remaining issues.
\end{workflowprompt}
\endgroup

\end{document}